\documentclass[]{opendatalab}
\usepackage{xcolor}
\usepackage{longtable}
\usepackage{listings}
\usepackage{graphicx}
\usepackage{caption}
\usepackage{subcaption}
\usepackage[numbers]{natbib}
\usepackage[table]{xcolor} 
\usepackage{booktabs}
\usepackage{graphicx}
\usepackage{newfloat}
\usepackage{threeparttable}
\usepackage{enumitem}
\definecolor{codegreen}{rgb}{0,0.6,0}
\definecolor{codegray}{rgb}{0.5,0.5,0.5}
\definecolor{codepurple}{rgb}{0.58,0,0.82}
\definecolor{backcolour}{rgb}{0.95,0.95,0.92}
\definecolor{promptcolor}{HTML}{D1D0F2}
\definecolor{promptcolorheader}{HTML}{bdbcec}
\definecolor{lightgray}{gray}{0.95}
\DeclareFloatingEnvironment{prompt}
\newcommand{\promptbox}[2]{
\begin{tcolorbox}[
top=0.3em,bottom=0.3em,left=0.5em,right=0.5em,
toptitle=0.3em,bottomtitle=0.2em,boxsep=0pt,
colframe=promptcolorheader,colback=promptcolor!50,boxrule=0.5pt,
]
\footnotesize
\end{tcolorbox}
}
\lstdefinestyle{mystyle}{
    backgroundcolor=\color{backcolour},   
    commentstyle=\color{codegreen},
    keywordstyle=\color{magenta},
    numberstyle=\tiny\color{codegray},
    stringstyle=\color{codepurple},
    basicstyle=\ttfamily\footnotesize,
    breakatwhitespace=false,         
    breaklines=true,                 
    captionpos=b,                    
    keepspaces=true,                 
    numbers=left,                    
    numbersep=5pt,                  
    showspaces=false,                
    showstringspaces=false,
    showtabs=false,                  
    tabsize=2
}

\title{Closing the Data Loop: Using OpenDataArena to Engineer Superior Training Datasets}

\author[1\dag]{Xin Gao}
\author[1\dag]{Xiaoyang Wang}
\author[1]{Yun Zhu}
\author[1]{Mengzhang Cai}
\author[1]{Conghui He}
\author[1\ast]{Lijun Wu}

\affiliation[1]{Shanghai Artificial Intelligence Laboratory, OpenDataLab, OpenDataArena}

\abstract{
The construction of Supervised Fine-Tuning (SFT) datasets is a critical yet under-theorized stage in the post-training of Large Language Models (LLMs), as prevalent practices often rely on heuristic aggregation without a systematic understanding of how individual samples contribute to model performance. In this report, we propose a paradigm shift from ad-hoc curation to a closed-loop dataset engineering framework using OpenDataArena (ODA), which leverages value-anchored rankings and multi-dimensional analysis to transform value benchmarking into feedback signals guiding dataset construction. We instantiate this methodology through two new datasets: \textbf{ODA-Math-460k}, a specialized mathematics reasoning dataset that utilizes a novel two-stage difficulty-aware pipeline to achieve State-of-the-Art (SOTA) results on benchmarks such as AIME and HMMT, and \textbf{ODA-Mixture (100k \& 500k)}, a series of multi-domain instruction datasets built via an ``Anchor-and-Patch'' strategy that outperforms significantly larger open-source baselines. Our empirical results demonstrate that ODA-driven datasets significantly improve both domain-specific reasoning and general utility while achieving superior data efficiency, validating a transition toward data-centric AI where transparent evaluation serves as the primary engine for engineering high-quality training data.
}

\metadata[Equal contribution]{Xin Gao, Xiaoyang Wang}
\correspondence{Lijun Wu, \email{wulijun@pjlab.org.cn}}
\metadata[OpenDataArena Project]{\url{https://opendataarena.github.io/}}
\metadata[ODA-Math \& ODA-Mixture]{\url{https://huggingface.co/datasets/OpenDataArena}}

\begin{document}

\vspace{-0.3cm}
\begin{figure}[th]
    \hspace{5cm}
    \includegraphics[width=0.3\linewidth]{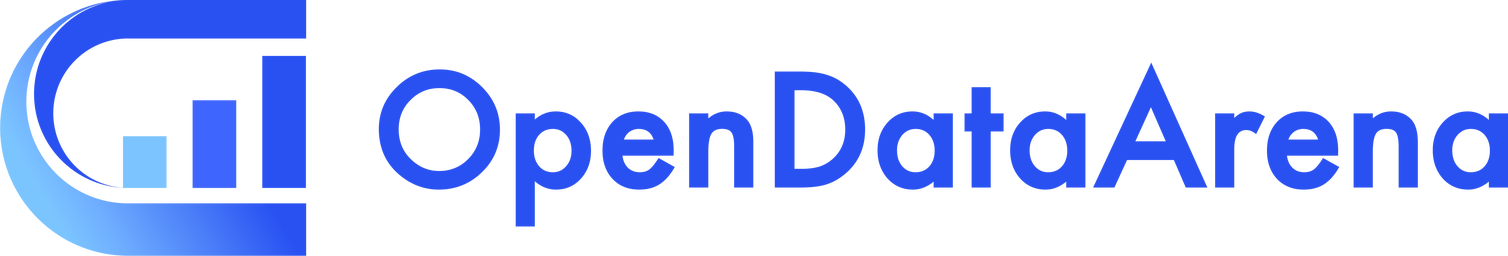}
    \label{fig:logo}
\end{figure}
\vspace{-0.2cm}

\maketitle

\vspace{-0.3cm}
\begin{figure}[!h]
    \centering
    \includegraphics[width=0.90\linewidth]{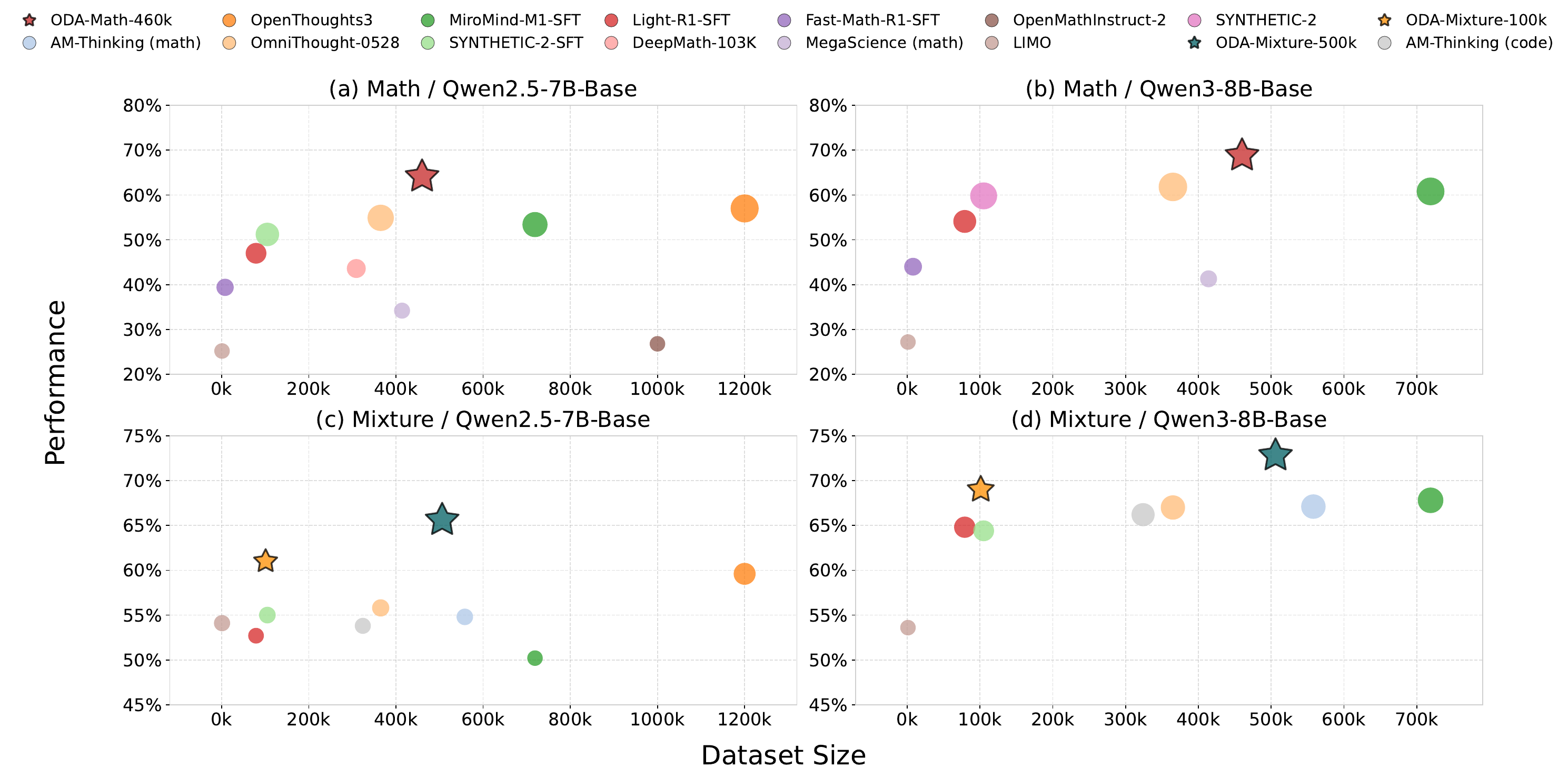}
    \caption{Performance vs. dataset size. ODA-Math/Mixture show great result with high efficiency.  
    }
    \
    \label{fig:brief_comparison}
\end{figure}
\vspace{-0.5cm}

\section{Introduction}
The post-training stage of Large Language Models (LLMs) is a critical determinant for unlocking instruction-following capabilities, complex reasoning, and general utility~\cite{zhao2025surveylargelanguagemodels, zhang2025instructiontuninglargelanguage, ouyang2022training, rafailov2024directpreferenceoptimizationlanguage}. Yet, despite its significance, the construction of Supervised Fine-Tuning (SFT) data mixtures remains under-theorized. While the open-source ecosystem offers a variety of candidate datasets across domains such as mathematics~\cite{numina-math, openmath-instruct2, openr1-math-220k}, code~\cite{liu2025rstar, ace-reason}, and general reasoning~\cite{open-thoughts, mega-science}, prevalent practices still rely heavily on heuristic aggregation. This approach—combining massive, heterogeneous datasets via intuition or trial-and-error—lacks a fundamental understanding of how individual sets contribute to downstream model capability~\cite{zhou2023lima, limo, muennighoff2025scalingdataconstrainedlanguagemodels}.

This opacity creates a substantial barrier to progress. The assessment of data utility is frequently confounded by inconsistent training recipes and evaluation variance~\cite{hochlehnert2025soberlookprogresslanguage, xu2024dposuperiorppollm}, as well as benchmark leakage that can artificially inflate performance metrics~\cite{xu2024benchmarkingbenchmarkleakagelarge, deng2024investigatingdatacontaminationmodern, shi2024detectingpretrainingdatalarge}. Without granular insight, practitioners are restricted to coarse-grained scaling strategies where the optimal composition for high-performance models remains stochastic and computationally inefficient~\cite{chen2024alpagasustrainingbetteralpaca}. To transcend this paradigm, it is imperative to shift from ad-hoc curation to quantitative data valuation. By rigorously measuring the marginal contribution of data samples, we seek to formalize SFT data construction as a systematic, metric-driven engineering discipline.

Our prior work, OpenDataArena (ODA)~\cite{cai2025opendataarenafairopenarena}, is a transparent platform originally designed for benchmarking data value. ODA offers comparative dataset benchmarking via the \href{https://opendataarena.github.io/leaderboard.html}{ODA-Leaderboard}, alongside a multi-dimensional data analysis system (\href{https://opendataarena-tool.readthedocs.io/en/latest/}{ODA-Tool}) by leveraging over 80 specialized data evaluators. In this work, we demonstrate that ODA serves not merely as a indicator for identifying relative performance, but as a robust source of supervisory signals for high-quality data construction. We argue that ODA’s value-anchored rankings and multiple data evaluators provide the feedback mechanism necessary to guide systematic dataset selection, composition, and verification. By operationalizing these rankings, we advance a methodology for developing training mixtures that are engineered for performance, traceability, and reproducibility.

\begin{figure}[!h]
    \centering
    \includegraphics[width=0.9\linewidth]{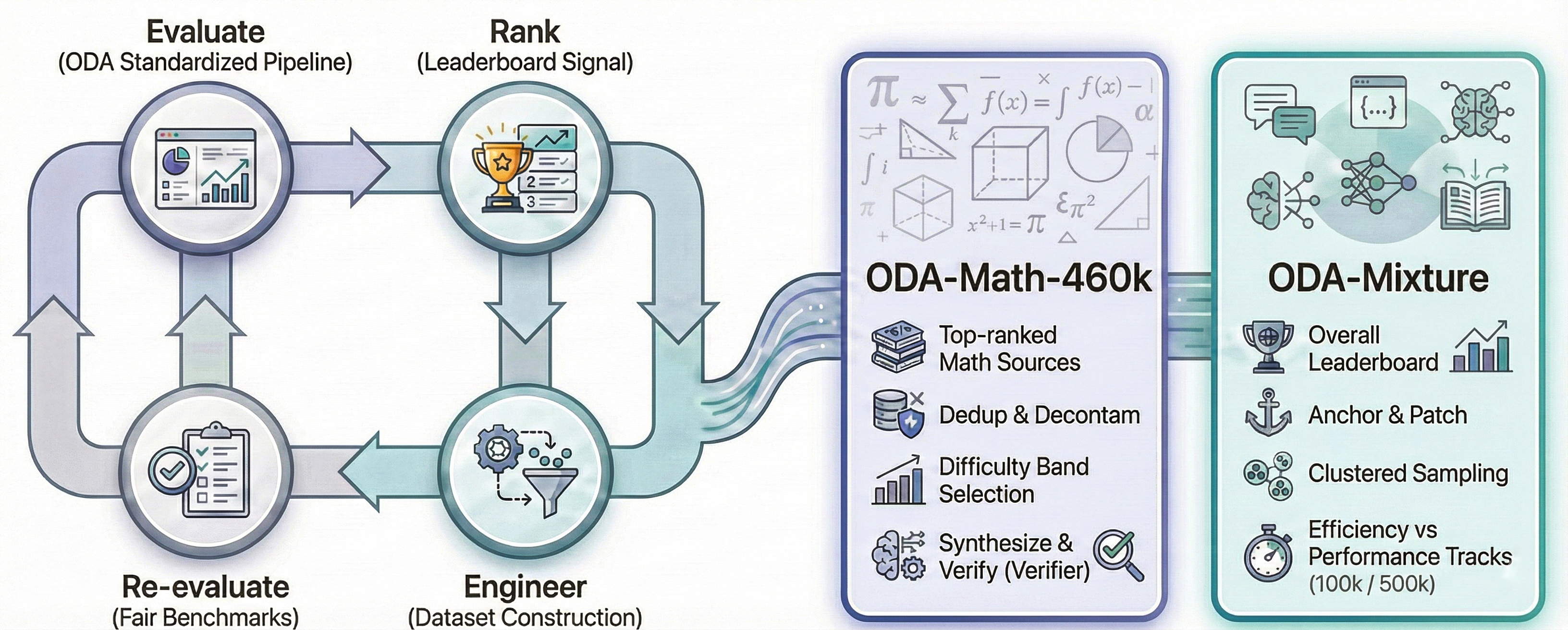}
    \caption{Overview of the construction of ODA-Math-460k and ODA-Mixture using OpenDataArena.}
    \label{fig:overall_framework}
\end{figure}

As shown in Figure~\ref{fig:overall_framework}, we transforms dataset construction from an intuition-driven craft into an iterative optimization process: \textbf{evaluate $\rightarrow$ rank $\rightarrow$ engineer $\rightarrow$ re-evaluate}. We instantiate this principle through two new SFT datasets engineered directly from ODA evidence, chosen to stress-test generality across both domain-specialized and broad-coverage settings. Specifically,

\textbf{(1) ODA-Math-460k: a math reasoning dataset engineered from top-ranked math sources.}
For basic quality control, we aggregate the most consistently effective open math datasets, then apply deduplication and benchmark decontamination to ensure fair evaluation and prevent leakage. To select most effective data: we introduce a novel two-stage selection pipeline—first filtering out problems that are too easy and then those that are too ambiguous or unsolvable—followed by a synthesize-and-verify distillation pipeline that yields high-quality step-by-step supervision with verifier-backed correctness. The resulting dataset emphasizes generalization to competition-style math while maintaining robustness across a broad math benchmark suite.

\textbf{(2) ODA-Mixture: a series of multi-domain instruction datasets engineered from the overall leaderboard.} 
To construct a dataset that performs strongly across multiple domains, such as general, math, code, and reasoning, we treat ODA’s overall leaderboard as a global signal and adopt an Anchor-and-Patch strategy: begin with an overall-strong, high-efficiency anchor dataset and patch it with domain-specialist datasets selected by ODA rankings. We further explore two regimes: an efficiency track that targets near-SOTA performance under a $\sim$100K sample budget, and a performance track that maximizes capability under a larger $\sim$500K budget. By combining ODA-guided source choice with diversity-aware sampling, we obtain mixtures that outperform prior datasets despite using substantially fewer samples.

This report makes three primary contributions to data-centric AI:

\begin{itemize}
    \item \textbf{A Closed-Loop Dataset Engineering Framework:} We propose a novel, ODA-driven methodology that utilizes leaderboard rankings and data evaluations as dynamic feedback signal for data selection. This allows for the iterative selection and composition of high-quality training corpora based on empirical performance.
    
    \item \textbf{High-Quality Open-Source Datasets:} We release ODA-Math-460k and ODA-Mixture-100k/500k, datasets constructed using ODA-guided data sourcing and difficulty-aware selection. These datasets serve as high-quality training sources for both domain-specialized reasoning and multi-domain generalization.
    
    \item \textbf{Empirical Validation of Data Efficiency:} Through comprehensive evaluation, we demonstrate that our ODA-driven datasets achieve SOTA performance with significantly improved data efficiency compared to strong open baselines. This validates ODA as a critical mechanism for driving data quality, not merely measuring it.
\end{itemize}

\section{ODA-Math-460k Dataset Construction}

\begin{figure}[!ht]
    \centering
    \includegraphics[width=0.9\linewidth]{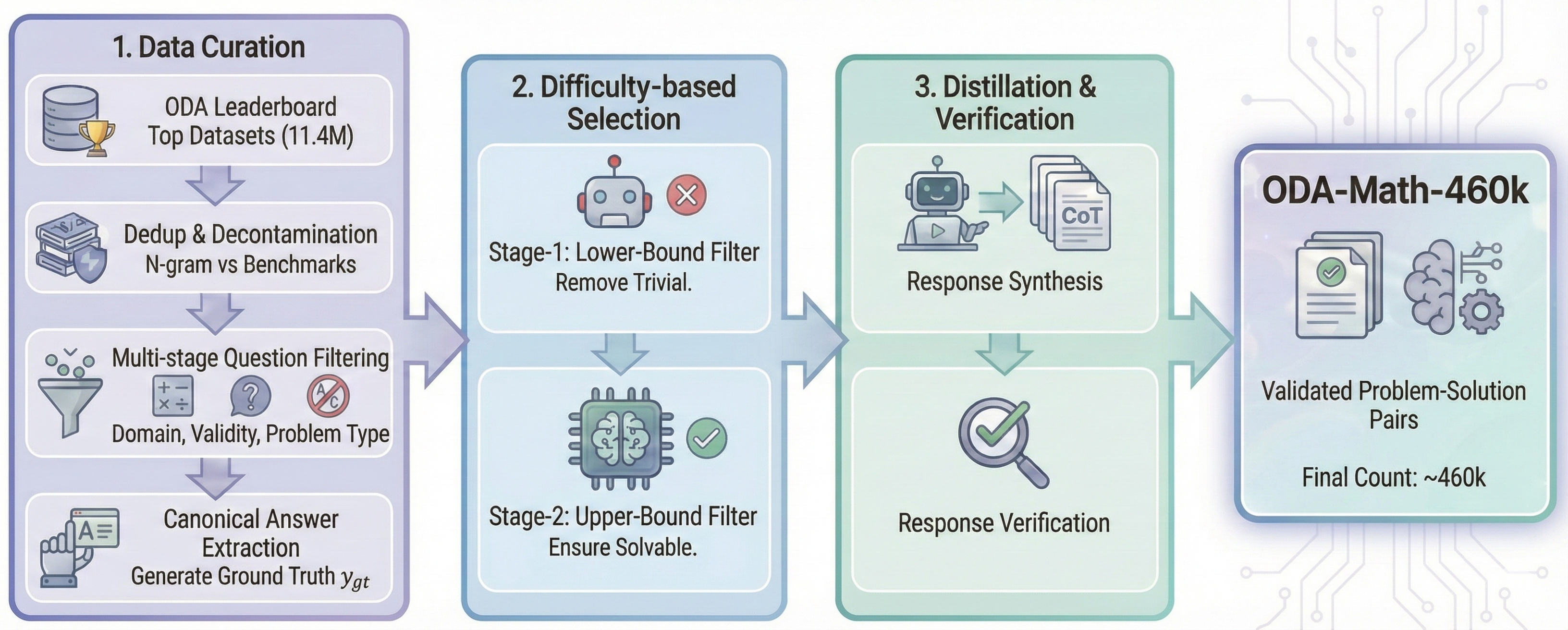}
    \caption{Overview of the construction of ODA-Math-460k.}
    \label{fig:overview-oda-math-460k}
\end{figure}

This section describes the end-to-end construction of ODA-Math-460k, a large-scale mathematics SFT corpus built via a `curate $\rightarrow$ select $\rightarrow$ distill $\rightarrow$ verify' pipeline. As shown in Figure~\ref{fig:overview-oda-math-460k}, we first curate an initial question pool by aggregating empirically effective math datasets from the ODA leaderboard~\cite{cai2025opendataarenafairopenarena}, followed by exact deduplication and $N$-gram decontamination against standard and recent competition-style benchmarks. We then apply multi-stage question filtering to ensure strict mathematical domain, well-posedness, and automatically verifiable problem types, and introduce an answer extraction step that converts solutions into canonical ground-truth answers while discarding instances with unusable solutions. Given the remaining large pool, we conduct a two-stage difficulty-based selection: a lower-bound filter removes problems solvable by a compact model, and an upper-bound filter removes unsolvable/ambiguous items by requiring solvability under a stronger reasoner. Finally, we distill high-quality reasoning traces using a teacher model and verify their correctness with a specialized verifier model, yielding the final training set consisting only of validated problem--solution pairs. The complete data statistics through the processing pipeline is shown in Table~\ref{tab:filtering_stats}.

\subsection{Data Curation}
We first employ a curation pipeline focused on basic quality, domain integrity, and verifiability. Following the aggregation of top-performing datasets, we perform rigorous deduplication and decontamination to remove redundancy and minimize benchmark leakage. The data is then refined through a multi-stage filtering process that enforces mathematical domain specificity, ensures structural validity, and excludes unverifiable problem types. Finally, an LLM-based extraction stage produces canonical ground-truth answers while discarding malformed solutions.

\subsubsection{Data Collection}
Leveraging the ODA leaderboard~\cite{cai2025opendataarenafairopenarena}, we identify and aggregate the top-ranking mathematics datasets that exhibit strong efficacy for the \texttt{Qwen} and \texttt{Llama} model families. Specifically, we select math-focused datasets that consistently yield competitive scores across standardized evaluation suites. The resulting collection serves as the initial question pool for ODA-Math-460k. The complete list of selected sources is provided in Table~\ref{tab:full_dataset_list} in Appendix. In this stage, we obtain a question pool of 11.4M. Besides, we create an independent dataset named MathLake\footnote{\url{https://huggingface.co/datasets/OpenDataArena/MathLake}}, which focuses on full coverage of questions and gathers data from over 50 math sources collected in ODA.

\subsubsection{Deduplication and Decontamination}
We first perform exact deduplication over all mathematical questions to eliminate redundant entries in the data pool. This step reduces overfitting to duplicated patterns and ensures more reliable estimates of data diversity.
Next, we conduct $N$-gram--based decontamination against a set of standard math benchmarks, including GSM8K~\cite{cobbe2021gsm8k}, Math500~\cite{hendrycks2021measuring}, Omni-Math~\cite{gao2024omni}, OlympiadBench-Math~\cite{he2024olympiadbench}, AIME'24~\cite{aime2025}, and AIME'25~\cite{aime2025}, as well as recent competition-style benchmarks such as CMIMC'25~\cite{cmimc2025}, HMMT-Feb'25~\cite{hmmt2025}, and BRUMO'25~\cite{brumo2025}. This process aims to minimize benchmark leakage and to prevent optimistic evaluation due to training–test overlap. Both deduplication and decontamination are implemented using the \texttt{NVIDIA Curator} toolkit~\cite{nvidia-nemo-curator}. After this stage, 62\% of the data are removed and the size of the problem set comes to 4.3M. 

\subsubsection{Question Filtering}
Following initial cleaning, we apply a multi-stage filtering pipeline to refine the dataset's domain specificity and quality.

\textbf{Domain Filtering.}
Although most source datasets are advertised as math-only, we observe a non-trivial fraction of out-of-domain questions (e.g., coding tasks, general science questions, or open-ended instruction-following data). To enforce a strictly mathematical domain, we apply LLM-based filtering to remove non-mathematical queries. Concretely, a classifier-style prompt instructs the model to decide whether a given instance is a genuine math problem. The full prompt and decision criteria are reported in Prompt~\ref{pmt:math_classification} in Appendix.

\textbf{Validity Filtering.}
We further identify ill-formed or unusable questions in the raw pool, including: 1) items presented as short phrases or declarative statements without a clearly posed question, and 2) questions with incomplete premises or missing references (e.g., referring to an unnamed theorem, a missing figure, or undefined notation). We employ an LLM-based validator, guided by a dedicated prompt (Prompt~\ref{pmt:math_validation} in Appendix), to filter out such instances. This step improves the syntactic and semantic well-posedness of the resulting dataset.

\textbf{Problem Type Filtering.}
After domain and validity filtering, we perform problem-type filtering to remove math questions whose answers are difficult to verify automatically. We adopt the \texttt{Big Math} toolkit~\cite{big-math} to identify and remove:
\begin{enumerate}[label=\arabic*)]
    \item Proof-based questions (due to evaluation ambiguity).
    \item Multiple-choice questions (to avoid performance inflation from random guessing).
    \item Binary (True/False) questions (also to avoid random guessing).
\end{enumerate}
The remaining pool predominantly consists of well-posed, free-form questions with objectively verifiable answers, which are better suited for large-scale supervised fine-tuning. The complete filtering stage further shrinks the problem set to 3.3M, which is 28.9\% of the original pool. 

\subsubsection{Solution Check and Final Answer Extraction}
\label{sec:answer-extraction}

The previous filtering stages operate purely on the question side, enforcing mathematical domain and suitable problem types. They do not, however, verify that the accompanying solutions provide a clean and extractable final answer. We therefore introduce an answer extraction stage that both constructs canonical ground-truth answers and filters out instances with unusable solutions.

Concretely, we provide the LLM with the original question and its accompanying solution and prompt it as a precise math answer extractor. The model is required to return exactly one XML-style tag \verb|<answer>...</answer>| containing only the final answer, without any intermediate reasoning. The extraction prompt encodes a small set of high-level rules: (1) prioritize the content of the final $\boxed{\cdot}$ expression when present; (2) otherwise, extract the last explicit numerical or symbolic result; and (3) handle multiple solutions and categorical answers in a consistent format. The full prompt is given in Prompt~\ref{pmt:answer_extraction} in Appendix.

We parse the \verb|<answer>| tag and apply post-processing to normalize the forms. If the extractor fails to produce a specific, concise answer (e.g., returns an empty tag or non-atomic text), we regard the original solution as invalid for our purposes and discard the corresponding problem–solution pair. The remaining instances yield canonical answers $y_{gt}$, which are then used as ground truth in later pass-rates-based problem selection stages and final distillation verification.

\begin{table}[t]
    \caption{Statistics of the data filtration pipeline. The process performs basic quality control and removes trivial and unsolvable problems to maximize training efficiency.}
    \centering
    \begin{tabular}{l c c r}
        \toprule
        \textbf{Pipeline Stage} & \textbf{Count} & \textbf{Percentage} & \textbf{Description} \\
        \midrule
        Raw Collection & 11.4M & 100\% & Initial aggregation \\
        Dedup \& Decontamination & 4.3M & 37.7\% & Match removal \\
        Question Filtering & 3.3M & 28.9\% & Invalid removal \\
        \midrule
        Stage-1 (Lower-Bound Filtering) & 815.3K & 7.2\% & Trivial removal (Pass@4$>$0) \\
        Stage-2 (Upper-Bound Filtering) & 459.6k & 4.0\% & Unsolvable removal (Pass@5$=$0) \\
        \bottomrule
    \end{tabular}
\label{tab:filtering_stats}
\end{table}

\subsection{Data Selection}

Given the large size of the curated question pool, we perform a two-stage selection procedure to retain only the most valuable problems for subsequent distillation. Intuitively, we aim to focus on questions that are (i) non-trivial for smaller models, yet (ii) solvable by a stronger reasoning model, thus forming a “learnable but challenging” band of difficulty. The selection is mainly based on the FailRateScorer\footnote{\url{https://github.com/OpenDataArena/OpenDataArena-Tool/tree/main/data_scorer/model_based}} data evaluator implemented by ODA-Tool.

\subsubsection{Stage-1: Lower-Bound Filtering}
We hypothesize that problems already solvable by a small model offer limited marginal training value. Therefore, the first stage removes questions that are reliably solved by a compact baseline. Concretely, we use \texttt{Qwen3-8B} to attempt each problem in a \emph{non-thinking} mode (i.e., direct generation without explicit chain-of-thought reasoning). For each problem $x$, we sample k=4 independent responses and compute a \textit{Pass-Rate} $\mathrm{Pass}@4(x)$ by matching the predicted final answer against the canonical ground-truth answer $y_{gt}$. We retain $x$ if and only if
\begin{equation}
    \mathrm{Pass}@4(x) = 0,
\end{equation}
i.e., none of the four attempts yields a correct solution. This stage filters out trivial problems that can be solved via memorization or shallow heuristics. The remaining problems come to 815.3K.

\subsubsection{Stage-2: Upper-Bound Filtering}
Among the remaining questions, some may be effectively unsolvable or overly ambiguous, which can hinder stable distillation. The second stage ensures that each retained problem is solvable by a stronger reasoning model. We employ \texttt{Qwen3-30B-A3B}~\cite{qwen3} in \emph{thinking} mode (i.e., decoding with detailed chain-of-thought reasoning), generating k=5 chain-of-thought traces for each problem $x$. 
Similarly, we then compute the \textit{Pass-Rate} score $\mathrm{Pass}@5(x)$ by comparing to $y_{gt}$ and keep only problems that satisfy
\begin{equation}
    \mathrm{Pass}@5(x) > 0.
\end{equation}
In other words, at least one of the six attempts must yield a correct solution. This upper-bound filtering removes over-difficult or ambiguous items and concentrates on problems that are both challenging and demonstrably solvable. We then have 459.6k problems left after removing unsolvable ones.

\subsection{Distillation and Verification}

\subsubsection{Response Synthesis}
To construct the final SFT dataset, we adopt a synthesize-and-verify paradigm to generate high-quality solutions for each selected problem. To balance computational cost and response quality, we choose the \texttt{AM-Thinking-v1} model~\cite{am-thinking-v1} as the teacher for response distillation. For each problem, we generate k=5 reasoning traces, each consisting of a step-by-step solution and a final answer. These candidate traces form a candidate set from which we later select verified responses.

\subsubsection{Response Verification}
We enforce the quality and correctness of the distilled responses using the \texttt{Compass-Verifier-7B} model~\cite{compass-verifier}. The verifier takes the problem statement, a generated response $y_{\mathrm{gen}}$, and the extracted ground-truth answer $y_{\mathrm{gt}}$ as input. It outputs a binary decision function $V(y_{\mathrm{gen}}, y_{\mathrm{gt}}) \in \{0,1\}$ indicating whether the generated solution is consistent with the reference answer. The final training corpus $D_{\mathrm{final}}$ is then defined as
\begin{equation}
    D_{\mathrm{final}} = \{(x, y_{\mathrm{gen}}) \mid V(y_{\mathrm{gen}}, y_{\mathrm{gt}}) = 1\},
\end{equation}  
i.e., the set of problem–solution pairs whose responses are successfully validated by the verifier. This procedure yields a high-quality SFT dataset with checked solutions and explicit reasoning traces. After verification, we have 459.6k problem–solution pairs in our final dataset $D_{\mathrm{final}}$. 

\subsection{Experiments}

\subsubsection{Experimental Setup}
\label{Experimental_Setup}
We evaluate the effectiveness of ODA-Math-460k through SFT for improving mathematical reasoning. We fine-tune two base models, \texttt{Qwen2.5-7B-Base} and \texttt{Qwen3-8B-Base}, on ODA-Math-460k and compare against representative open math SFT datasets under matched evaluation protocols. Training is conducted with the LlamaFactory\footnote{\url{https://github.com/hiyouga/LLaMA-Factory}} toolkit, and inference-time evaluation is implemented with OpenCompass~\cite{opencompass} and vLLM~\cite{vllm}.

\paragraph{Training configuration.}
All SFT runs use identical optimization hyperparameters unless otherwise specified. We use DeepSpeed ZeRO-3 with configuration ds\_z3\_config, the default template, and a long-context cutoff length of 32,768 tokens. We enable packed sequences to improve throughput. The per-device batch size is 2 with gradient accumulation steps of 2, yielding an effective batch size of 4 sequences per device. We train for 3 epochs with a peak learning rate of $5\times 10^{-5}$, a cosine learning-rate schedule, and a warmup ratio of 0.1. We additionally enable use\_liger\_kernel for faster training. A complete summary is provided in Table~\ref{tab:training-hyperparameters}.

\paragraph{Inference configuration.}
We evaluate with vLLM using model-specific inference presets (Table~\ref{tab:inference-hyperparameters}). For \texttt{Qwen2.5-7B-Base}, we use greedy decoding (temperature=0). For \texttt{Qwen3-8B-Base}, we use stochastic decoding with temperature=0.6, top-p=0.95, and top-k=20. For both models, we set max-out-len=32,768 and use vLLM with cutoff to respect context length constraints.

\paragraph{Benchmarks and evaluation protocol.}
We evaluate on a broad suite of math benchmarks spanning grade-school arithmetic to competition-level reasoning: GSM8K~\cite{cobbe2021gsm8k}, Math-500~\cite{hendrycks2021measuring}, Omni-Math~\cite{gao2024omni}, OlympiadBench~\cite{he2024olympiadbench}, AIME'24~\cite{aime2025}, AIME'25~\cite{aime2025}, CMIMC'25~\cite{cmimc2025}, HMMT-Feb'25~\cite{hmmt2025}, and BRUMO'25~\cite{brumo2025}. Table~\ref{tab:math-eval-benchmarks} details the evaluator used for each benchmark and the reported metric. For standard datasets (GSM8K, Math-500, and OlympiadBench), we use \texttt{xVerify-9B-C}~\cite{xVerify} as the answer judge and report accuracy. Omni-Math is evaluated with \texttt{Omni-Judge}~\cite{gao2024omni}. For contest-style benchmarks (AIME'25, HMMT-Feb'25, CMIMC'25, BRUMO'25), we use \texttt{CompassVerifier-7B}~\cite{compass-verifier}. For AIME'24 and all contest-style tasks, we report the average accuracy over 8 independent runs to reduce variance from sampling and judge sensitivity, following Table~\ref{tab:math-eval-benchmarks}.

\subsubsection{Main Results}
\newcommand{\rot}[1]{\rotatebox{45}{\textbf{\small{#1}}}}

\begin{table*}[ht]
\caption{Performance comparison. We evaluate models across various mathematical benchmarks. The best scores are highlighted in \textbf{bold}, and the second-best scores are \underline{underlined}. ODA-Math-460k achieves competitive performance across multiple metrics.}
\centering
\begin{threeparttable}
\resizebox{\textwidth}{!}{%
\begin{tabular}{l | r | c @{\hspace{1pt}} c @{\hspace{1pt}} c @{\hspace{1pt}} c @{\hspace{1pt}} c @{\hspace{1pt}} c @{\hspace{1pt}} c @{\hspace{1pt}} c @{\hspace{1pt}} c |c}
\toprule
\textbf{Dataset} & \textbf{Size} & 
\rot{GSM8K} & 
\rot{Math500} & 
\rot{Omni-Math} & 
\rot{Olympiad*} & 
\rot{AIME'24} & 
\rot{AIME'25} & 
\rot{CMIMC'25} & 
\rot{HMMT'25} & 
\rot{BRUMO'25} & 
\textbf{AVG} \\
\midrule
\multicolumn{12}{c}{\textbf{Qwen2.5-7B-Base}} \\
\midrule
Qwen2.5-7B-Base~\cite{qwen2-5} &  & 80.0 & 50.2 & 26.0 & 35.9 & 6.7 & 6.7 & 10.0 & 0.0 & 20.0 & 26.2 \\
\midrule
LIMO~\cite{limo} & 817 & 92.1 & 66.8 & 21.6 & 34.9 & 4.6 & 1.7 & 0.0 & 0.0 & 5.4 & 25.2 \\
OpenMathInstruct-2~\cite{openmath-instruct2} & 1M & 91.6 & 65.9 & 22.5 & 30.7 & 6.7 & 5.0 & 5.0 & 0.0 & 13.6 & 26.8 \\
MegaScience (math)~\cite{mega-science} & 414k & 90.1 & 77.8 & 28.7 & 44.5 & 16.7 & 15.0 & 8.1 & 0.0 & 26.7 & 34.2 \\
Fast-Math-R1-SFT~\cite{fast-math-r1} & 8k & 90.6 & 80.0 & 35.8 & 50.3 & 23.3 & 26.7 & 7.5 & 8.3 & 31.7 & 39.4 \\
DeepMath-103K~\cite{deep-math-103k} & 309k & 92.1 & 92.0 & 45.4 & 60.2 & 34.2 & 31.7 & 10.0 & 11.7 & 15.0 & 43.6 \\
Light-R1-SFT~\cite{light-r1} & 79k & 92.0 & 88.0 & 43.3 & 60.2 & 38.3 & 26.7 & 22.5 & 13.3 & 38.3 & 47.0 \\
SYNTHETIC-2-SFT~\cite{synthetic-2} & 105k & 92.1 & 90.0 & 54.5 & 67.4 & 45.0 & 35.0 & 19.7 & 20.0 & 36.7 & 51.2 \\
MiroMind-M1-SFT~\cite{miro-mind} & 719k & \underline{93.9} & 91.6 & 48.1 & 66.3 & 55.0 & 30.0 & 27.5 & 18.3 & 50.0 & 53.4 \\
OmniThought-0528~\cite{omni-thought} & 365k & 93.2 & 89.8 & 54.3 & 68.1 & 50.4 & 40.0 & 25.0 & 28.3 & 45.0 & 54.9 \\
OpenThoughts3~\cite{open-thoughts} & 1.2M & 91.7 & 93.8 & 44.8 & 68.8 & \underline{60.0} & 45.0 & 27.5 & 31.7 & 50.0 & 57.0 \\
AM-Thinking (math)~\cite{am-thinking-v1-distilled} & 558k & 92.9 & \textbf{96.2} & \underline{60.6} & \textbf{74.2} & \textbf{63.3} & \underline{50.0} & \underline{27.8} & \underline{36.7} & \textbf{63.3} & \underline{62.8} \\
\rowcolor{lightgray} ODA-Math-460k & 460k & \textbf{94.3} & \underline{95.4} & \textbf{62.6} & \underline{70.9} & 56.7 & \textbf{56.7} & \textbf{35.0} & \textbf{45.0} & \underline{60.0} & \textbf{64.1} \\
\midrule
\multicolumn{12}{c}{\textbf{Qwen3-8B-Base}} \\
\midrule
Qwen3-8B-Base~\cite{qwen3} & & 92.0 & 79.6 & 30.6 & 47.2 & 6.7 & 10.8 & 4.7 & 0.0 & 16.7 & 32.0 \\
\midrule
LIMO~\cite{limo} & 817 & 83.9 & 69.0 & 21.8 & 31.3 & 12.5 & 8.8 & 2.2 & 1.7 & 13.8 & 27.2 \\
MegaScience (math)~\cite{mega-science} & 414k & 93.4 & 84.8 & 35.8 & 57.6 & 25.4 & 17.9 & 11.3 & 12.1 & 33.8 & 41.3 \\
Fast-Math-R-SFT~\cite{fast-math-r1} & 8k & 92.8 & 86.6 & 39.6 & 61.0 & 28.8 & 25.8 & 14.1 & 13.3 & 34.2 & 44.0 \\
Light-R1-SFT~\cite{light-r1} & 79k & 93.8 & 92.6 & 48.5 & 69.7 & 54.6 & 31.3 & 22.8 & 25.0 & 48.8 & 54.1 \\
SYNTHETIC-2-SFT~\cite{synthetic-2} & 105k & 93.9 & 93.8 & 58.8 & 71.5 & 58.8 & 45.8 & 28.4 & 32.9 & 54.2 & 59.8 \\
MiroMind-M1-SFT~\cite{miro-mind} & 719k & \underline{94.8} & \textbf{96.8} & 54.5 & \underline{77.0} & 62.9 & 47.5 & 25.6 & 27.5 & 60.4 & 60.8 \\
OmniThought-0528~\cite{omni-thought} & 365k & 94.2 & 95.4 & 59.0 & 74.9 & \textbf{67.9} & 45.4 & 31.3 & 35.8 & 52.5 & 61.8 \\
AM-Thinking (math)~\cite{am-thinking-v1-distilled} & 558k & \textbf{95.2} & 95.6 & \underline{64.5} & \textbf{77.5} & 65.8 & \underline{54.6} & \underline{36.3} & \underline{41.3} & \underline{62.5} & \underline{65.9} \\
\rowcolor{lightgray} ODA-Math-460k & 460k & 94.3 & \underline{96.0} & \textbf{66.9} & 76.3 & \textbf{67.9} & \textbf{63.3} & \textbf{41.6} & \textbf{45.4} & \textbf{67.5} & \textbf{68.8} \\
\bottomrule
\end{tabular}
}
\begin{tablenotes}[flushleft] 
    \small \item * Note: Olympiad refers to the English math subset of the original OlympiadBench datasets.
\end{tablenotes}
\end{threeparttable}
\label{tab:performance_comparison}
\end{table*}





Table~\ref{tab:performance_comparison} presents a comprehensive performance evaluation of ODA-Math-460k against several SOTA mathematical SFT datasets. Our dataset demonstrates superior performance across both the Qwen2.5-7B and Qwen3-8B architectures, achieving the highest average scores in both categories (64.1\% and 68.8\%, respectively). Besides the SOTA performance scores, we also have the following observations:
\begin{itemize}
    \item \textbf{Efficiency and scaling.} A key takeaway from these results is the high information density of ODA-Math-460k. When compared to datasets of significantly larger scale, such as OpenMathInstruct-2 (1M samples) and OpenThoughts-3 (1.2M samples), ODA-Math-460k achieves superior results while utilizing less than half the training data. This suggests that our two-stage selection process—which aggressively filters out both trivial items and unsolvable noise—successfully bypasses the ``diminishing returns'' often seen in massive, uncurated distilled corpora.
    \item \textbf{Mastery of competition-level reasoning.} The most significant performance gaps are observed in high-difficulty, competition-style benchmarks. ODA-Math-460k shows a distinct advantage on the 2025 iterations of AIME, CMIMC, and HMMT. Notably: 1) On HMMT'25, ODA-Math-460k outperforms the nearest competitor (AM-Thinking) by 8.3\% on Qwen2.5-7B and 4.1\% on Qwen3. 2) On CMIMC'25, our dataset establishes a new SOTA for these base models, achieving a score of 41.6\% on Qwen3-8B.
\end{itemize}

The superiority in these specific benchmarks—which require long-horizon reasoning and are less likely to be contaminated due to their recent release—indicates that our dataset fosters genuine reasoning robustness rather than pattern matching.

We attribute this performance to two primary factors. (1) \textbf{Optimal difficulty band.} By focusing on the ``frontier'' of the model's capability through 2-stage selection, we ensure the training signal is neither too simple nor too complex. (2) \textbf{Verified trace quality.} Unlike datasets that rely solely on model-generated outputs, our use of verifier-backed distillation ensures that the reasoning paths (CoT) leading to the correct answer are logically sound.
In summary, ODA-Math-460k provides a highly efficient recipe for mathematical SFT, proving that rigorous data quality and difficulty alignment are more impactful than raw dataset volume.

\subsubsection{Investigating Experiments}

To validate the efficacy of our data pipeline, we conduct investigative experiments using \texttt{Qwen2.5-7B-Base} as the backbone model. These experiments study three critical dimensions: query sampling strategies, the impact of problem type filtration, and the necessity of answer verification.

\paragraph{Comparison of data selection strategies.}
We evaluated six distinct sampling strategies, each selecting 100k samples from the initial query pool via the ODA-Tool\footnote{\url{https://opendataarena-tool.readthedocs.io/en/latest/}}~\cite{opendataarena_tool_2025}:
\begin{enumerate}[label=\arabic*)]
    \item \textit{Random}: Uniform random sampling from the query pool, serving as the baseline.
    \item \textit{Diversity}: Queries are mapped to an embedding space and clustered; samples are then drawn from each cluster to ensure broad semantic and topical coverage.
    \item \textit{\href{https://opendataarena-tool.readthedocs.io/en/latest/model-based-evaluation/\#deitacscorer}{Deita-Complexity}}: A model-based scorer~\cite{deita} that ranks queries by complexity (1--6). We selected the 100k queries with the highest complexity scores.
    \item \textit{\href{https://opendataarena-tool.readthedocs.io/en/latest/model-based-evaluation/\#answerprobscorer}{Answer-Prob}}: This approach~\cite{answerprob} assesses query-answer quality by calculating the conditional probability $P(\text{Answer} \mid \text{Query})$ relative to the marginal $P(\text{Answer})$.
    \item \textit{\href{https://opendataarena-tool.readthedocs.io/en/latest/model-based-evaluation/\#thinkingprobscorer}{Thinking-Prob}}: A model-based scorer~\cite{thinkprob} that estimates the likelihood of a model engaging in explicit reasoning chains, reflecting perceived cognitive demand.
    \item \textit{Pass-Rate}: Queries are sampled based on the pass rate across four inference trials using \texttt{Qwen3-8B}. We prioritized queries with a 0 pass rate.
\end{enumerate}

As summarized in Table~\ref{tab:math-benchmarks}, the \textit{Pass-Rate} strategy significantly outperformed all alternatives, achieving a 44.7\% average performance. These results suggest that targeting a model's ``knowledge frontier''---specifically queries where it consistently fails---provides a more potent training signal than general complexity or diversity metrics. Notably, while \textit{Deita-Complexity} and \textit{Diversity} offered marginal gains over the baseline, \textit{Answer-Prob} and \textit{Thinking-Prob} resulted in performance degradation. Consequently, we identify explicit \textit{Pass-Rate} as a more robust proxy for curriculum selection in mathematical reasoning than traditional model-based scoring.

\begin{table}[ht]
\centering
\begin{threeparttable}
\caption{Evaluation of different sampling strategies (100k samples) on \texttt{Qwen2.5-7B-Base}.}
\label{tab:math-benchmarks}
\begin{tabular}{l|cccc|c}
\toprule
\textbf{Strategy} & \textbf{Omni-Math*} & \textbf{Olympiad} & \textbf{AIME'24} & \textbf{AIME'25} & \textbf{AVG} \\ \midrule
Random & 22.5 & 59.8 & 31.7 & \textbf{35.0} & 37.2 \\
Diversity & 19.9 & 60.4 & 40.0 & 33.3 & 38.4 \\
Deita-Complexity~\cite{deita} & 21.7 & 61.9 & 40.0 & 33.3 & 39.2 \\
Answer-Prob~\cite{answerprob} & 19.9 & 57.9 & 26.7 & 23.3 & 31.9 \\
Thinking-Prob~\cite{thinkprob} & 8.7 & 49.0 & 15.0 & 23.3 & 24.0 \\
Pass-Rate & \textbf{28.7} & \textbf{66.0} & \textbf{50.0} & 34.2 & \textbf{44.7} \\ \bottomrule
\end{tabular}
\begin{tablenotes}[flushleft] 
    \small \item * Note: We sample 500 questions from original Omni-Math benchmark for experiment efficiency.
\end{tablenotes}
\end{threeparttable}
\end{table}

\paragraph{Impact of problem type filtration.} 
We investigate the impact of problem type on SFT performance by identifying a subset of queries at the highest Art of Problem Solving (AoPS) difficulty level (10). This subset is characterized by a high proportion of proof-based problems, which often lack easily verifiable numerical answers. By comparing 20k samples of unfiltered and filtered queries, we observe that excluding poorly posed or non-verifiable proof problems significantly improves outcomes. As shown in Table~\ref{tab:problem_type_filtration}, the model demonstrates superior generalization when trained on well-posed questions with deterministic, verifiable solutions.

\begin{table}[h]
\centering
\caption{SFT performance comparison (20k samples) with and without problem type filtration.}
\label{tab:problem_type_filtration}
\begin{tabular}{l|cccc|c}
\toprule
\textbf{Setting} & \textbf{GSM8k} & \textbf{Math500} & \textbf{Omni-Math} & \textbf{Olympiad} & \textbf{AVG} \\ 
\midrule
Unfiltered & 85.6 & 81.6 & 43.1 & 49.9 & 65.1 \\ 
Filtered & \textbf{87.7} & \textbf{82.0} & \textbf{45.1} & \textbf{50.9} & \textbf{66.4} \\
\bottomrule
\end{tabular}
\end{table}

\paragraph{Impact of answer verification.} 
To determine the necessity of answer verification for distilled CoT traces, we compare the SFT performance of 100k queries using verified versus unverified responses. As illustrated in Table~\ref{tab:response_verification}, training on verified CoT traces consistently yields superior results across all benchmarks. Most notably, verification led to a 3.8\% absolute improvement on the Olympiad benchmark and a 2.0\% increase in the overall average. These findings underscore that ensuring the correctness of distilled reasoning paths is critical for high-quality mathematical SFT.

\begin{table}[h]
\centering
\caption{SFT performance comparison (100k samples) on verified vs. unverified distilled responses.}
\label{tab:response_verification}
\begin{tabular}{l|cccc|c}
\toprule
\textbf{Setting} & \textbf{GSM8k} & \textbf{Math500} & \textbf{Omni-Math} & \textbf{Olympiad} & \textbf{AVG} \\ 
\midrule
Unverified & 91.7 & 90.2 & 55.8 & 62.2 & 75.0 \\ 
Verified & \textbf{91.8} & \textbf{92.6} & \textbf{57.5} & \textbf{66.0} & \textbf{77.0} \\
\bottomrule
\end{tabular}
\end{table}

\subsection{Data Analysis}

\begin{figure}[!h]
    \centering
    \includegraphics[width=0.8\linewidth]{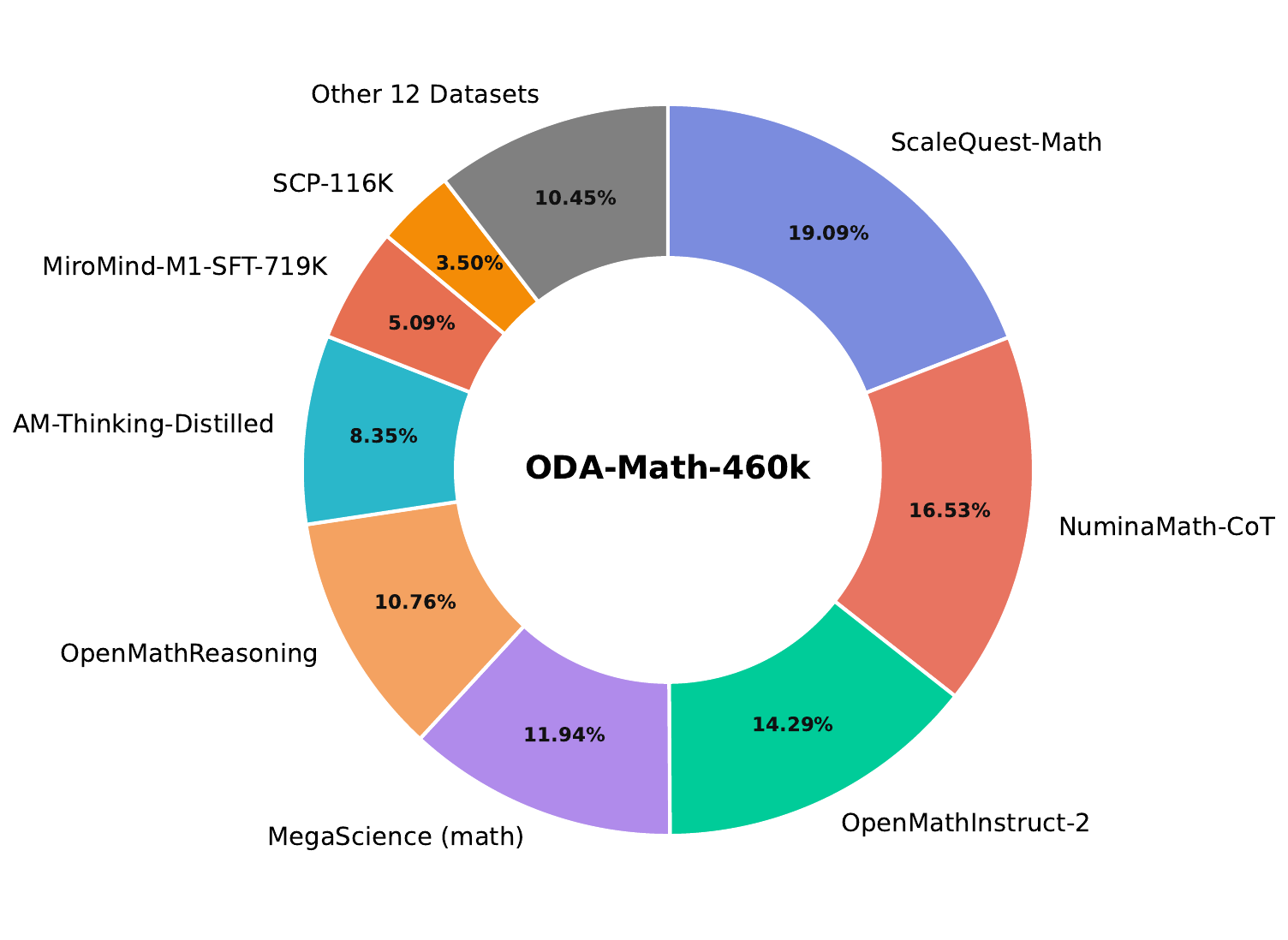}
    \caption{Top problem contributors for ODA-Math-460k.}
    \label{fig:math_source_dist}
\end{figure}

\paragraph{Source composition.} 
ODA-Math-460k is constructed by aggregating high-quality problems from a diverse suite of existing mathematical datasets. After the data pipeline, the final dataset consists of 20 data sources and the top contributors are shown in Figure~\ref{fig:math_source_dist}. Major contributors—specifically ScaleQuest-Math~\cite{scale-quest-math}, NuminaMath-CoT~\cite{numina-math}, OpenMathInstruct-2~\cite{openmath-instruct2}, MegaScience (math)~\cite{mega-science}, and OpenMathReasoning~\cite{open-math-reasoning}—each account for approximately 10–20\% of the total volume. The remaining mass is comprised of a "long tail" of supplementary sources, including AM-Thinking-Distilled~\cite{am-thinking-v1-distilled}, MiroMind-M1-SFT-719K~\cite{miro-mind}, SCP-116K~\cite{scp-116k} and other 12 datasets. Detailed source distribution can be found in Table~\ref{tab:source_dist_booktabs} in Appendix. This balanced mixture prevents the dataset from being dominated by any single distribution; consequently, ODA-Math-460k inherits the collective strengths of multiple curated sources while smoothing out their individual biases and idiosyncrasies. Notably, although top contributors such as ScaleQuest-Math and OpenMathInstruct-2 contain high proportions of synthetic data and do not independently achieve top-tier rankings on the ODA leaderboard, our rigorous selection and distillation pipeline ensures that their latent value is fully extracted and utilized.

\begin{figure}[!h]
    \centering
    \includegraphics[width=0.95\linewidth]{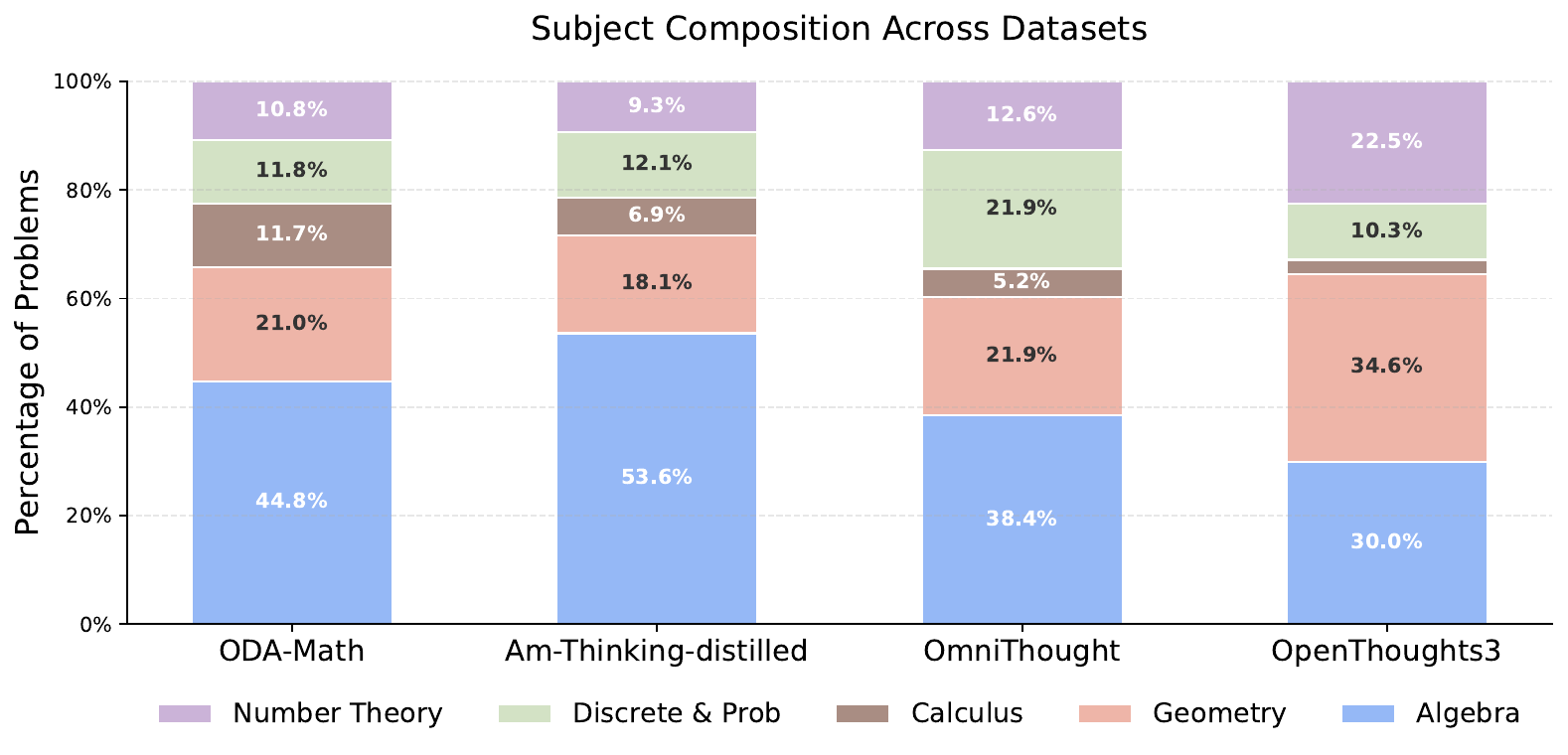}
    \caption{Comparison on subjects distribution of problems across datasets.}
    \label{fig:math_subject_distribution}
\end{figure}

\paragraph{Problem category distribution.} 
As illustrated in Figure~\ref{fig:math_subject_distribution}, ODA-Math-460k provides a more balanced subject composition than other datasets. While all datasets are algebra-heavy, Am-Thinking-distilled over-concentrates on algebra (53.6\%) with relatively few geometry or calculus problems, and OpenThoughts3 strongly emphasizes geometry and number theory with almost no calculus. In contrast, ODA-Math-460k keeps substantial algebra (44.8\%) but allocates roughly 20–22\% to geometry and around 11\% each to calculus, discrete mathematics \& probability, and number theory. This mitigates subject bias and better approximates the heterogeneous composition of real-world math workloads and evaluation suites, which mix algebraic manipulation, geometric reasoning, combinatorics, and continuous mathematics. As a result, models trained on ODA-Math-460k are less likely to exhibit sharp performance drops on underrepresented topics, improving both average and worst-case accuracy.

\begin{figure}[!h]
    \centering
    \includegraphics[width=0.95\linewidth]{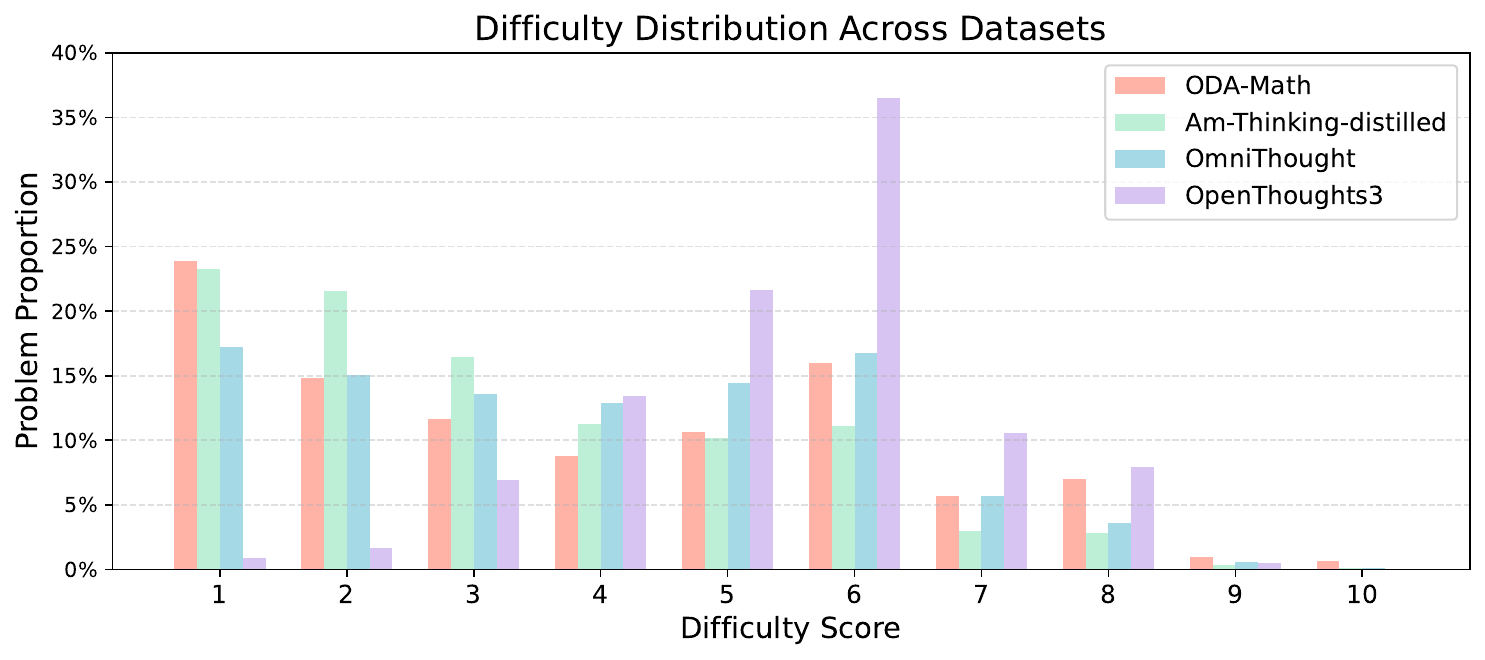}
    \caption{Comparison on difficulty distribution of problems across datasets. The scoring standard can be found in Prompt~\ref{pmt:math_difficulty} in Appendix.}
    \label{fig:math_difficulty_distribution}
\end{figure}

\begin{figure}[!h]
    \centering
    \includegraphics[width=0.8\linewidth]{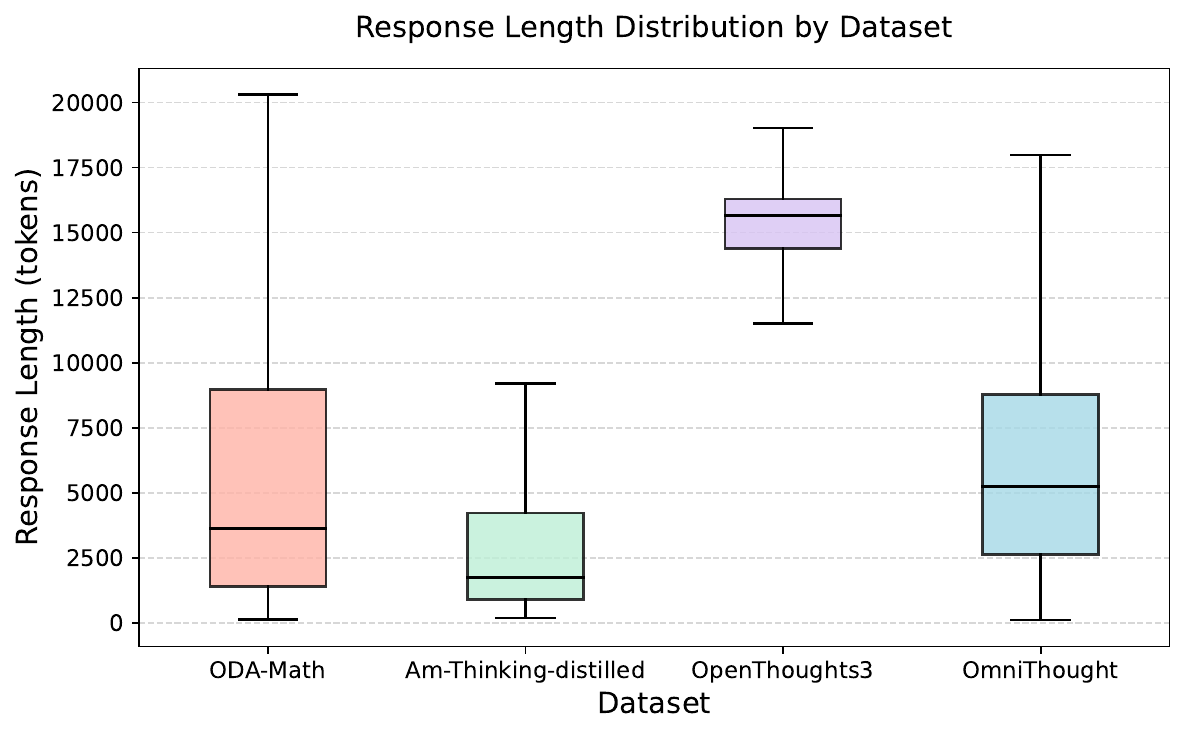}
    \caption{Comparison on response length distribution across datasets.}
    \label{fig:math_length_distribution}
\end{figure}

\paragraph{Problem difficulty distribution.} 
Building on this balanced subject coverage, following~\cite{guha2025openthoughts,yang2025select2reason}, we next use an LLM-as-Judge way to study difficulty scores of each question, based on which we analyze the difficulty distribution in Figure~\ref{fig:math_difficulty_distribution}. The difficulty profile of ODA-Math-460k resembles a well-designed curriculum more closely than competing datasets. By contrast, Am-Thinking-distilled and OmniThought are skewed toward easier items, while OpenThoughts3 concentrates most of its mass near difficulty 6 with very few genuinely easy or very hard problems. For SFT, such a balanced spectrum is crucial: easier items stabilize training and teach basic patterns, mid-range items match the difficulty of most public benchmarks, and harder items encourage models to learn longer, more global reasoning strategies. This curriculum-like distribution likely helps models learn robust behaviors across the full difficulty range, rather than overfitting to a narrow band.

\paragraph{Response length.} 
The response-length statistics in Figure~\ref{fig:math_length_distribution} show that ODA-Math-460k strikes a favorable balance between brevity and depth of reasoning supervision. Its solutions are markedly longer and more varied than those in Am-Thinking-distilled, providing rich step-by-step rationales instead of short pattern-matching answers, yet they remain substantially shorter than the extremely long chains in OpenThoughts3 and less uniformly verbose than those in OmniThought. This diversity of lengths teaches models to modulate explanation depth according to problem complexity while keeping token budgets manageable for SFT. Together with the balanced difficulty and subject distributions discussed above, this calibrated reasoning signal likely yields cleaner gradients, better utilization of context, and improved generalization, which in turn explains why ODA-Math-460k leads to SOTA SFT performance.


\section{ODA-Mixture Dataset Construction}
This section details the construction of ODA-Mixture. We first outline the data curation and decontamination protocols used to ensure evaluation integrity. Next, we describe the specific sampling strategies tailored for the efficiency and performance tracks. Finally, we provide a holistic evaluation of the ODA-Mixture, using both quantitative experiments and qualitative analyses to verify our design choices and performance gains.


\subsection{Data Curation}

\begin{figure}
    \centering
    \includegraphics[width=0.9\linewidth, trim=0 100 0 50]{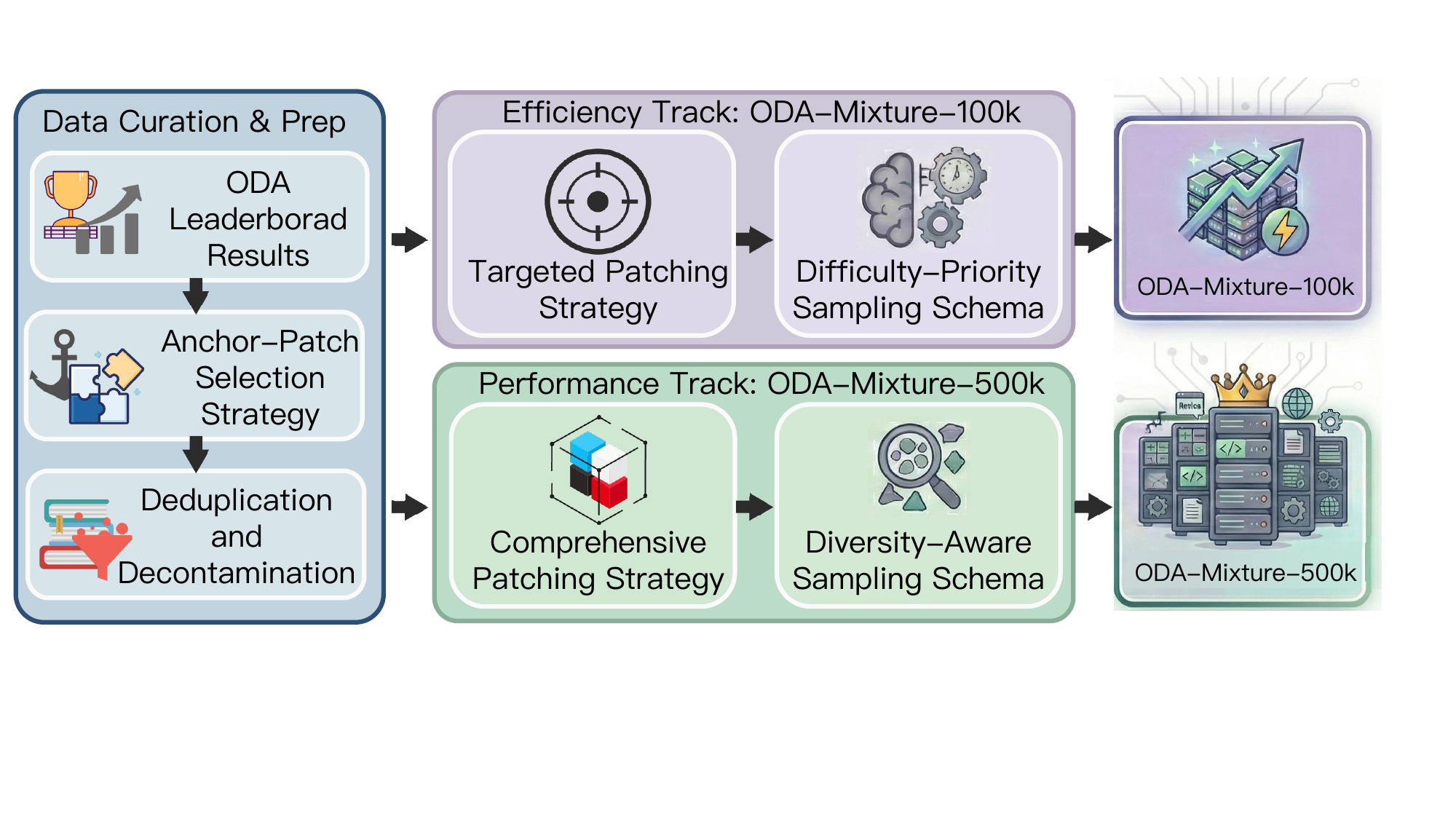}
    \caption{Overview of the construction of ODA-Mixture-500k and ODA-Mixture-100k.}
    \label{fig:ODA-MIX-Pipeline}
\end{figure}

\subsubsection{Data Collection}

The data collection process leverages the ODA leaderboard~\cite{cai2025opendataarenafairopenarena}, utilizing its objective benchmarks to guide sourcing. Specifically, we use the results of the Qwen2.5-7B-Base model on the ODA leaderboard as our primary reference to ensure a consistent and reliable quality signal. To construct a dataset that achieves robust performance across all domains, we adopt an ``Anchor-and-Patch'' strategy guided by two core criteria: absolute leaderboard performance and data efficiency---defined as the performance improvement a dataset yields per unit of data. By prioritizing sources with high information density, we aim to establish a compact yet potent foundation for subsequent operations.

Following these principles, we first establish a performance anchor by selecting LIMO~\cite{limo}. Despite containing only 817 samples, LIMO exhibits the highest data efficiency on the Overall leaderboard, serving as our foundational ``quality compass.'' Recognizing that a single compact anchor may have specialized gaps, we then form a candidate pool by patching the collection with complementary datasets that excel in specific ODA in-domain leaderboards. For the Math domain, we select AM-Thinking-v1-Distilled-math~\cite{am-thinking-v1-distilled} (Rank 1). For Code, we incorporate AM-Thinking-v1-Distilled-code~\cite{am-thinking-v1-distilled}(Rank 3). The General domain is addressed using math-gpt-4o-200k~\cite{math-gpt4o-200k} (Rank 2), and for Reasoning, we utilize SYNTHETIC-2-SFT-verified~\cite{synthetic-2} (Rank 1). This structured selection ensures the final ODA-Mixture maintains high information density while achieving comprehensive coverage across diverse tasks.

\subsubsection{Deduplication and Decontamination}
Since our candidate pool—formed by the anchor and patch datasets—consists of top-tier datasets from the ODA leaderboard, they inherently possess a baseline level of quality control. Consequently, our pre-processing focus shifted to strict rigorous fairness. We perform standard exact-match deduplication to remove redundancy. Furthermore, to ensure the integrity of our evaluation, we conduct comprehensive $n$-gram decontamination against the test sets of the 18 benchmarks used in our experiments, eliminating any risk of test data leakage.

\subsection{Data Selection}
Starting from the curated candidate pool, we derive two specialized versions of ODA-Mixture to explore different data-scaling frontiers. The first, ODA-Mixture-100k, serves as an efficiency track designed to achieve high-quality results within a highly constrained data budget. The second, ODA-Mixture-500k, represents a performance track aimed at maximizing overall model capability using a more extensive data volume.

\subsubsection{The Efficiency Track: ODA-Mixture-100k}
The objective of this track is to develop a high-density dataset that maintains strong performance across the ODA benchmarks using a minimal viable data volume on the order of $\sim$100k samples.
For this track, we utilize the LIMO anchor and patch it specifically with the Math and Code sources. This targeted patching strategy is driven by our observation that while LIMO provides a robust foundation for general instruction following, its specialized reasoning capabilities in formal logic and algorithmic tasks remain the primary bottleneck. By concentrating our limited sample budget on these two high-impact domains rather than diluting it across the entire candidate pool, we can more effectively bridge the anchor's capability gaps while preserving extreme data efficiency. 

Since these domains consist predominantly of high-difficulty reasoning benchmarks, the patching data must focus on strengthening these challenging capabilities. Under a constrained budget, prioritizing harder examples is more effective than maximizing surface-level diversity. Prior work~\cite{zhao2024long,open-thoughts,yang2025select2reason} has shown that sequence length correlates with problem difficulty in reasoning-heavy settings. Guided by this observation, we first generate embeddings for all candidate samples in the Math and Code pools and apply K-Means clustering to ensure broad semantic coverage. Within each cluster, we then apply a difficulty-priority sampling scheme, preferentially selecting instances with higher estimated difficulty—operationalized via longer token counts—until the sampling budget is exhausted. This approach ensures that ODA-Mixture-100k provides sufficient exposure to complex, multi-step reasoning cases despite its compact scale.

\subsubsection{The Performance Track: ODA-Mixture-500k}
The objective of this track is to prioritize the upper bound of model performance across all evaluated domains, utilizing a more generous data budget of approximately 500k samples.
With the increased data budget, the optimization focus shifts from targeted gap-filling to comprehensive distributional coverage. Unlike the efficiency track, we patch the LIMO anchor using all four major domains—Math, Code, General, and Reasoning—to ensure a balanced and versatile intelligence profile. At this scale, further gains depend less on concentrating exclusively on hard examples and more on capturing a broad spectrum of skills, reasoning patterns, and instruction formats. 

To this end, we adopt a diversity-aware clustered sampling strategy. We first generate embeddings for all candidate samples across the four domain sources and apply K-Means clustering to obtain a balanced semantic partition. Then, within each cluster, we perform random sampling without explicit difficulty bias. This design avoids over-concentration on long-tail complex instances and instead promotes wide semantic and stylistic coverage, which is critical for improving performance uniformly across diverse benchmarks from different domains. By utilizing the full breadth of the candidate pool, ODA-Mixture-500k aims to provide a robust and well-rounded training signal that scales effectively with the larger data volume.
\begin{table*}[t!]
\centering
\caption{Leaderboard performance comparison on Overall four domain benchmarks. Best results within each section are shown in \textbf{bold}, and second-best results are \underline{underlined}. Eff. denotes Data Efficiency.}
\begin{tabular}{l | c c | c c c c | c}
\toprule
\textbf{Dataset} & \textbf{Size} & \textbf{Eff.} & \textbf{General} & \textbf{Math} & \textbf{Code} & \textbf{Reasoning} & \textbf{Avg} \\
\midrule
\multicolumn{8}{c}{\textbf{Qwen2.5-7B-Base}} \\
\midrule
Qwen2.5-7B-Base~\cite{qwen2-5}                         &     &               & 51.4          & 39.8          & 50.1          & 42.7          & 46.0 \\
\midrule

MiroMind-M1-SFT~\cite{miro-mind} & 719k & +0.006& 52.0 & 71.0 & 26.3 & 51.5 & 50.2 \\
Light-R1-SFT~\cite{light-r1} & 79k & +0.084 & 55.5 & 64.4 & 38.8 & 51.9 & 52.7 \\
AM-Thinking (code)~\cite{am-thinking-v1-distilled} & 324k & +0.024	& 49.9 & 52.3 & \textbf{68.7} & 44.4 & 53.8 \\

LIMO~\cite{limo}      & 817  & \textbf{+9.920} & \underline{60.7} & 44.0       & 57.9 & 53.8          & 54.1 \\
AM-Thinking (math)~\cite{am-thinking-v1-distilled} & 558k & +0.016         & 57.7          & \textbf{77.4} & 39.5          & 44.8          & 54.8 \\
SYNTHETIC-2-SFT~\cite{synthetic-2}       & 105k & +0.086 & 51.3      & 69.8          & 40.1          & \underline{58.9} & 55.0 \\
OmniThought-0528~\cite{omni-thought}              & 365k & +0.027         & 47.1          & 71.2          & 47.6          & 57.2          & 55.8 \\
OpenThoughts3~\cite{{open-thoughts}}          & 1.2M & +0.011         & 45.5          & 71.8          & \underline{67.0} & 54.3          & 59.6 \\

\rowcolor{lightgray} ODA-Mixture-100k & 101k & \underline{+0.149} & 56.8 & 71.2 & 64.4 & 51.5 & \underline{61.0}\\
\rowcolor{lightgray} ODA-Mixture-500k & 506k & +0.039  & \textbf{63.4}  & \underline{72.8}  & 66.7  & \textbf{59.6} & \textbf{65.6} \\

\midrule
\multicolumn{8}{c}{\textbf{Qwen3-8B-Base}} \\
\midrule
Qwen3-8B-Base~\cite{qwen3}  &   &   & 58.7  & 51.2  & 52.4   & 50.6   & 53.2 \\
\midrule
LIMO~\cite{limo}      & 817  & \textbf{+0.490} & 61.7 & 46.0 & 52.7 & 54.1 & 53.6 \\
SYNTHETIC-2-SFT~\cite{synthetic-2}       & 105k & +0.107 & 59.5 & 75.4 & 56.1 & \underline{66.6} & 64.4 \\
Light-R1-SFT~\cite{light-r1} & 79k & +0.168 & 64.9 & 71.8 & 59.0 & 63.6 & 64.8 \\
AM-Thinking (code)~\cite{am-thinking-v1-distilled} & 324k & +0.045	& 64.8 & 64.9 & \textbf{75.8} & 59.3 & 66.2 \\
OmniThought-0528~\cite{omni-thought} & 365k & +0.043	& 55.8 & \underline{78.3} & 68.1 & 66.0 & 67.0 \\
AM-Thinking (math)~\cite{am-thinking-v1-distilled} & 558k & +0.028	& \underline{65.9} & \textbf{79.7} & 59.5 & 63.2 & 67.1 \\
MiroMind-M1-SFT~\cite{miro-mind} & 719k & +0.023& 64.5 & 77.2 & 63.6 & 65.8 & 67.8 \\

\rowcolor{lightgray} ODA-Mixture-100k & 101k & \underline{+0.177} & 61.1	& 77.3 & \underline{73.2} & 64.7 & \underline{69.0} \\
\rowcolor{lightgray} ODA-Mixture-500k & 506k & +0.042 & \textbf{71.2} & 77.2 & 73.0 & \textbf{69.7} & \textbf{72.8} \\

\bottomrule
\end{tabular}
\label{tab:overall_leaderboard_comparison}
\end{table*}

\begin{table*}[t!]
\centering
\caption{Leaderboard performance comparison on the General domain benchmarks. For each backbone, we include the \textbf{top-5 datasets} from the corresponding ODA leaderboards for comparison. Best results within each section are shown in \textbf{bold}, and second-best results are \underline{underlined}. Eff. denotes Data Efficiency.}
\resizebox{\textwidth}{!}{%
\begin{tabular}{l | c c | c c c c | c}
\toprule

\textbf{Dataset} & \textbf{Size} & \textbf{Eff.} & \textbf{DROP} & \textbf{IFEVAL} & \textbf{AGIEVAL} & \textbf{MMLU PRO} & \textbf{Avg} \\
\midrule
\multicolumn{8}{c}{\textbf{Qwen2.5-7B-Base}} \\
\midrule
Qwen2.5-7B-Base~\cite{qwen2-5}  &     &    & 68.3          & 35.5          & 57.7          & 44.2          & 51.4 \\
\midrule

tulu-3-sft-personas-algebra~\cite{lambert2024tulu}      & 20k  & \textbf{+0.482}         & 76.9          & 48.9          & 65.6          & 52.9          & 61.1 \\
Magpie-Reasoning-V1~\cite{magpie}       & 150k & \underline{+0.065}         & 78.4          & \underline{55.5} & 63.3       & 47.4          & 61.1 \\
MegaScience ~\cite{mega-science}                 & 1.25M& +0.008         & 75.9          & 48.1          & 65.3          & \textbf{57.7} & 61.8 \\
math-gpt-4o-200k ~\cite{math-gpt4o-200k} & 200k & +0.055         & 74.8          & 52.4          & \textbf{68.8} & 53.8          & 62.5 \\
TextbookReasoning~\cite{mega-science}            & 652k & +0.018         & \underline{79.7}          & 49.2          & \underline{67.7} & \underline{54.8}    & \underline{62.9} \\
\rowcolor{lightgray}ODA-Mixture-100k                 & 101k & +0.053 & \textbf{87.3} & 39.1 & 61.2 & 39.6  & 56.8 \\
\rowcolor{lightgray}ODA-Mixture-500k                 & 506k & +0.024 & 75.2 & \textbf{57.7} & 66.0 & \underline{54.8} & \textbf{63.4} \\

\midrule
\multicolumn{8}{c}{\textbf{Qwen3-8B-Base}} \\
\midrule
Qwen3-8B-Base                    &     &              & 71.5          & 45.9          & 61.1          & 56.2          & 58.7 \\
\midrule
MiroMind-M1-SFT  ~\cite{miro-mind}  & 719k & +0.008 & 85.0          & 43.5          & 74.2          & 55.1          & 64.5 \\
AM-Thinking (code) ~\cite{am-thinking-v1-distilled}   & 324k & +0.019 & 88.2          & 50.1          & 70.4          & 50.8          & 64.9 \\
Light-R1-SFT  ~\cite{light-r1}   & 79k  & \underline{+0.078} & 83.4          & 46.8          & 71.6          & \underline{57.7} & 64.9 \\
AM-Thinking (math)~\cite{am-thinking-v1-distilled}    & 558k & +0.013 & \textbf{93.3} & 38.7    & \underline{74.7} & 56.9       & 65.9 \\
Raiden-DeepSeek-R1 ~\cite{sequelbox_raiden_deepseek_r1}            & 63k & \textbf{+0.143} & 82.3 & \underline{63.2} & 71.9 & 53.3 & \underline{67.7} \\

\rowcolor{lightgray}ODA-Mixture-100k    & 101k & +0.024 & \underline{91.6} & 44.2 & 73.3 & 35.1 & 61.1 \\
\rowcolor{lightgray}ODA-Mixture-500k   & 506k & +0.025 & 77.4 & \textbf{71.2} & \textbf{75.3} & \textbf{60.8} & \textbf{71.2} \\
\bottomrule
\end{tabular}
}

\label{tab:general_leaderboard_comparison}
\end{table*}

\begin{table*}[t!]
\centering
\caption{Leaderboard performance comparison on the Math domain benchmarks. For each backbone, we include the \textbf{top-5 datasets} from the corresponding ODA leaderboards for comparison. Best results within each section are shown in \textbf{bold}, and second-best results are \underline{underlined}. Eff. denotes Data Efficiency; OMNI and OLYMP denote the Omni-Math and Olympiad benchmarks, respectively.}
\
\resizebox{\textwidth}{!}{%

\begin{tabular}{l | c c | c c c c c | c}
\toprule

\textbf{Dataset} & \textbf{Size} & \textbf{Eff.} & \textbf{GSM8K} & \textbf{MATH500} & \textbf{OMNI} & \textbf{OLYMP} & \textbf{AIME'24} & \textbf{Avg} \\
\midrule

\multicolumn{9}{c}{\textbf{Qwen2.5-7B-Base}} \\
\midrule
Qwen2.5-7B-Base  ~\cite{qwen2-5}   &     &               & 80.0          & 50.2          & 26.0          & 35.9          & 6.7           & 39.8 \\
\midrule
SYNTHETIC-2-SFT~\cite{synthetic-2} & 105k & \underline{+0.286}& 92.1         & 90.0          & 54.5 & 67.4       & 45.0          & 69.8 \\
MiroMind-M1-SFT~\cite{miro-mind} & 719k & +0.043   & \textbf{93.9} & 91.6          & 48.1          & 66.3          & 55.0          & 71.0 \\
OmniThought-0528~\cite{omni-thought}     & 365k & +0.086 & \underline{93.2} & 89.8    & 54.3  & 68.1   & 50.4          & 71.2 \\
OpenThoughts3~\cite{open-thoughts}    & 1.2M & +0.027  & 91.7  & \underline{93.8} & 44.8 & 68.8 & \underline{60.0} & 71.8 \\
AM-Thinking (math)~\cite{am-thinking-v1-distilled} & 558k & +0.067 & 92.9 & \textbf{96.2} & \textbf{60.6} & \textbf{74.2} & \textbf{63.3} &\textbf{77.4} \\

\rowcolor{lightgray}ODA-Mixture-100k & 101k & \textbf{+0.311} & 90.8 & 91.0 & \underline{55.8} & 68.6 & 50.0 & 71.2\\
\rowcolor{lightgray}ODA-Mixture-500k & 506k & +0.065 & 92.9 & 92.0 & 52.8 & \underline{69.7} & 56.7 & \underline{72.8} \\
\midrule

\multicolumn{9}{c}{\textbf{Qwen3-8B-Base}} \\
\midrule
Qwen3-8B-Base   &     &   & 92.0	&79.6	&30.6&	47.2&	6.7&	51.2 \\
\midrule
Light-R1-SFT ~\cite{light-r1}& 79k & \textbf{+0.261} & 93.8	&92.6 & 48.5 &69.7 &54.6	&71.8\\
SYNTHETIC-2-SFT~\cite{synthetic-2}& 105k & +0.230 & 94.0 &93.8 & 58.8 & 71.5 &58.8 &75.4 \\
MiroMind-M1-SFT~\cite{miro-mind} & 719k & +0.036  & \underline{94.8} & \textbf{96.8}&54.5	&\underline{77.0}	&62.9 &77.2 \\
OmniThought-0528 ~\cite{omni-thought} & 365k &+0.074  & 94.2	&95.4	&59.0	&75.0	&\textbf{67.9} & \underline{78.3} \\
AM-Thinking (math)~\cite{am-thinking-v1-distilled} & 558k & +0.051 &\textbf{95.2} &\underline{95.6}&\textbf{64.5}&\textbf{77.5}&\underline{65.8}&\textbf{79.7} \\

\rowcolor{lightgray}ODA-Mixture-100k & 101k & \underline{+0.258} & 93.8	& 94.2 & \underline{61.2} & 73.3 &64.1 & 77.3\\
\rowcolor{lightgray}ODA-Mixture-500k & 506k & +0.051 & 94.7	& 94.0 & 59.8 & 75.7 & 62.1 & 77.2 \\

\bottomrule
\end{tabular}
}
\label{tab:math_leaderboard_comparison_full}
\end{table*}

\begin{table*}[t!]
\centering
\caption{Leaderboard performance comparison on the Code domain benchmarks. For each backbone, we include the \textbf{top-5 datasets} from the corresponding ODA leaderboards for comparison. Best results within each section are shown in \textbf{bold}, and second-best results are \underline{underlined}. Eff. denotes Data Efficiency.}
\begin{tabular}{l | c c | c c c c | c}
\toprule

\textbf{Dataset} & \textbf{Size} & \textbf{Eff.} & \textbf{HumanEval} & \textbf{MBPP} & \textbf{LCB} & \textbf{HumanEval+} & \textbf{Avg} \\
\midrule

\multicolumn{8}{c}{\textbf{Qwen2.5-7B-Base}} \\
\midrule
Qwen2.5-7B-Base  ~\cite{qwen2-5}                     &     &               & 77.4          & 71.6          & 8.2           & 43.3          & 50.1 \\
\midrule
EpiCoder-func~\cite{wang2025epicoder}    & 380k & +0.031         & 82.1          & 73.4          & 15.9          & 76.4          & 62.0 \\
OpenThoughts3~\cite{open-thoughts}    & 1.2M & +0.014         & 78.7          & \textbf{79.4} & 31.9          & 78.1          & 67.0 \\
AM-Thinking (code)~\cite{am-thinking-v1-distilled} & 324k & \underline{+0.057}     & 78.1     & \underline{79.0} & 40.9       & 76.8          & 68.7 \\
OpenCodeReasoning~\cite{ahmad2025opencodereasoning}     & 752k & +0.029         & \textbf{88.4} & 76.7    & \underline{42.7} & \textbf{79.3} & \underline{71.7} \\
rStar-Coder~\cite{liu2025rstar}         & 990k & +0.022    & \underline{86.6} & 77.4  &\textbf{46.2} & \underline{78.7} & \textbf{72.2} \\

\rowcolor{lightgray}ODA-Mixture-100k & 101k & \textbf{+0.141} & 81.1 & 75.9 & 28.0 & 72.6 & 64.4 \\
\rowcolor{lightgray}ODA-Mixture-500k & 506k & +0.033 & 80.5 & 78.6 & 33.3 & 74.4 & 66.7 \\

\midrule

\multicolumn{8}{c}{\textbf{Qwen3-8B-Base}} \\
\midrule
Qwen3-8B-Base  ~\cite{qwen3}  &     &               & 82.9          & 75.5 & 16.9          & 34.2          & 52.4 \\
\midrule
MegaScience~\cite{mega-science}    & 1.25M & +0.008   & 83.5	&69.7	&21.9	&75.0	&62.5 \\
OpenThoughts-114k~\cite{open-thoughts} & 114k & \underline{+0.097} &72.0	&75.5	&31.5	&75.0	&63.5 \\
MiroMind-M1-SFT~\cite{miro-mind} & 719k & +0.016 & 82.3 & 79.8 &21.2	&71.3 & 63.6 \\
OmniThought-0528~\cite{omni-thought}   & 365k & +0.043    & \textbf{91.5}	&\underline{86.8} &29.4&	64.6	&68.1 \\
AM-Thinking (code)~\cite{am-thinking-v1-distilled} & 324k & +0.073 &\textbf{91.5} &\textbf{89.5}	&\textbf{43.0}	&\underline{79.3}&	\textbf{75.8} \\

\rowcolor{lightgray}ODA-Mixture-100k & 101k & \textbf{+0.206} & \underline{87.2}	&85.6	&38.7  &\textbf{81.1} & \underline{73.2} \\
\rowcolor{lightgray}ODA-Mixture-500k & 506k & +0.041 & \textbf{91.5} &83.3	&\underline{41.0}&76.8	&73.0 \\

\bottomrule
\end{tabular}

\label{tab:code_leaderboard_comparison}
\end{table*}

\begin{table*}[t!]
\centering
\caption{Leaderboard performance comparison on the Reasoning domain benchmarks. For each backbone, we include the \textbf{top-5 datasets} from the corresponding ODA leaderboards for comparison. Best results within each section are shown in \textbf{bold}, and second-best results are \underline{underlined}. Eff. denotes Data Efficiency. KOR denotes KOR-BENCH.}

\begin{tabular}{l | c c | c c c c c | c}
\toprule

\textbf{Dataset} & \textbf{Size} & \textbf{Eff.} & \textbf{ARC-C} & \textbf{BBH} & \textbf{GPQA} & \textbf{CALM} & \textbf{KOR} & \textbf{Avg} \\
\midrule

\multicolumn{9}{c}{\textbf{Qwen2.5-7B-Base}} \\
\midrule
Qwen2.5-7B-Base~\cite{qwen2-5}  &     &    & 36.6  & 69.5   & 34.9 & 39.2 & 33.3       & 42.7 \\
\midrule
OpenO1-SFT~\cite{xia2025generative}         & 78k& \underline{+0.146} & 80.7   & 72.5    & \textbf{33.8} & \underline{36.0}       & 47.2          & 54.0 \\
OpenThoughts3~\cite{open-thoughts}    & 1.2M & +0.010         & 82.7          & 70.7          & 28.8          & 32.9          & 56.2          & 54.3 \\
Orca-agentinstruct~\cite{mitra2024agentinstruct} & 1.2M & +0.010       & 86.8          & 70.2          & 31.3          & \textbf{38.6}          & 48.5          & 55.1 \\
OmniThought-0528~\cite{omni-thought}     & 365k & +0.040  & \underline{90.5} & 76.8 & 30.8  & 29.1 & \textbf{58.9} & 57.2 \\
SYNTHETIC-2-SFT~\cite{synthetic-2} & 105k & \textbf{+0.154} & \textbf{93.2} & \underline{81.1} & \underline{31.8} & 29.5 & \underline{58.8} & \underline{58.9} \\

\rowcolor{lightgray}ODA-Mixture-100k & 101k & +0.087 & 84.1 & 68.1 & 22.7 & 33.2 & 49.4 & 51.5 \\
\rowcolor{lightgray}ODA-Mixture-500k & 506k & +0.033 & 88.1 & \textbf{82.7} & 31.3 & 35.0 & \underline{58.8} & \textbf{59.6} \\

\midrule

\multicolumn{9}{c}{\textbf{Qwen3-8B-Base}} \\
\midrule
Qwen3-8B-Base~\cite{qwen3}   &     &     & 37.3    & 78.1 & 41.9  & 48.9 & 46.6          & 50.6 \\
\midrule
OpenThoughts-114k~\cite{open-thoughts}   & 114k & +0.036 & 91.2 & 82.3 & 42.4  & 43.1 & 57.8  & 63.3 \\
Light-R1-SFT ~\cite{light-r1} & 79k& \underline{+0.069}& 92.2	& 84.5 & 46.5 & 34.2 & 60.4	& 63.6 \\
MiroMind-M1-SFT~\cite{miro-mind} & 719k & +0.007  & \underline{92.9}   & 86.5    & 47.5   & 40.2   & 62.1  & 65.8 \\
OmniThought-0528~\cite{omni-thought}     & 365k & +0.019  & \textbf{93.9} & \underline{86.8} & \underline{51.0}       & 33.4    & \underline{65.0}  & 66.0 \\
SYNTHETIC-2-SFT~\cite{synthetic-2} & 105k & \underline{+0.069} & \underline{92.9}    & 86.6  & \textbf{55.6} & 34.0   & 64.2   & \underline{66.6} \\

\rowcolor{lightgray}ODA-Mixture-100k & 101k & \textbf{+0.087}& 91.5 & 82.4 & 45.0 & \underline{45.0} & 59.4 & 64.7 \\
\rowcolor{lightgray}ODA-Mixture-500k & 506k & +0.038 &92.2 & \textbf{88.0} & 48.0 & \textbf{54.8} & \textbf{65.8} & \textbf{69.7} \\

\bottomrule
\end{tabular}
\label{tab:reasoning_leaderboard_comparison}
\end{table*}

\subsection{Experiments}

\subsubsection{Experimental Setup}
All models are fine-tuned using the same backbone and training configuration as ODA-Math (see Section \ref{Experimental_Setup}). Both the training and evaluation protocols strictly follow the standardized ODA setup to ensure fair and controlled comparisons across all datasets. 

Evaluation is conducted on the full ODA benchmark suite, covering four major domains.  
For the General domain, we report results on DROP~\cite{dua2019drop}, IFEVAL~\cite{zhou2023instruction}, AGIEVAL~\cite{zhong2024agieval}, and MMLU-Pro~\cite{wang2024mmlu}.  
For the Math domain, we evaluate on GSM8K~\cite{cobbe2021gsm8k}, MATH500~\cite{hendrycks2021measuring}, Omni-Math~\cite{gao2024omni}, OlympiadBench~\cite{he2024olympiadbench}, and AIME2024~\cite{aime2025}.  
For the Code domain, we include HumanEval~\cite{chen2021evaluating}, MBPP~\cite{austin2021program}, LCB (V5)~\cite{jain2024livecodebench}, and HumanEval+~\cite{liu2023your}.  
For the Reasoning domain, evaluation is performed on ARC-C~\cite{clark2018think}, BBH~\cite{suzgun2022challenging}, CALM~\cite{chen2024causal}, and KOR-BENCH~\cite{ma2024kor}.
More detailed evaluation settings are provided in Tab.\ref{tab:math-eval-benchmarks}.

\subsubsection{Main Results}

Tables~\ref{tab:overall_leaderboard_comparison}--\ref{tab:reasoning_leaderboard_comparison} summarize the performance of ODA-Mixtures on the Overall leaderboard and its four constituent domains: General, Math, Code, and Reasoning, evaluated on both \texttt{Qwen2.5-7B} and \texttt{Qwen3-8B} backbones, we assess these datasets through absolute scores and Data Efficiency, a metric defined in Appendix~\ref{sec:Appendix_Data_Efficiency} representing performance gain per unit of post-training data. From the results, we can observe three primary findings:
\begin{itemize}
    \item \textbf{Setting new SOTA performance.} The ODA-Mixture-500k sets new SOTA on the overall benchmarks across both backbones. Specifically, on \texttt{Qwen2.5-7B}, it surpasses the previous SOTA dataset, OpenThoughts3-1.2M, by nearly 6 points while using less than half training samples. On \texttt{Qwen3-8B}, it exceeds the prior best-performing data, MiroMind-M1-SFT-719K, by nearly 5 points with a reduction of 200k samples. These significant margins demonstrate that ODA-guided selection enables highly effective scaling without the need for excessive data volume.
    \item \textbf{Extreme efficiency.} Targeting the extreme efficiency track, ODA-Mixture-100k achieves remarkable results with a minimal data budget. Despite using only $\sim$100k samples, it outperforms previous SOTA datasets by 1.4 points on \texttt{Qwen2.5-7B} and 1.2 points on \texttt{Qwen3-8B}. Notably, it attains the highest Data Efficiency among all evaluated datasets, confirming that ODA can identify high-quality samples that deliver disproportionate performance gains. The stability of ODA mixtures in surpassing prior SOTA across different backbones further underscores the robustness of our platform.
    \item \textbf{Balanced capability and domain robustness.} A closer examination of sub-domain performance further illustrates these trends. In the General and Reasoning domains, both ODA-Mixtures consistently rank among the top-tier candidates across both backbones, suggesting that our construction method naturally yields broad coverage without explicit domain-specific optimization. In high-difficulty domains such as Math and Code, we observe backbone-dependent behavior: while the mixtures remain competitive on \texttt{Qwen2.5-7B}, they rival or even surpass specialized domain-specific datasets when paired with \texttt{Qwen3-8B}. Interestingly, ODA-Mixture-100k occasionally achieve similar performance with its 500k counterpart. We attribute this to the stronger intrinsic capabilities of \texttt{Qwen3} and the efficacy of difficulty-focused data in rapidly amplifying those strengths under tight data budgets.
\end{itemize}



These results support the central hypothesis of ODA: leveraging unified, domain-aware signals for data selection produces efficient, general-purpose mixtures that enhance performance across diverse domains while maintaining superior data efficiency without complex heuristic tuning.

\subsubsection{Ablation Studies}
\label{Ablation_Studies}
We conduct ablation studies on \texttt{Qwen2.5-7B} to analyze key design choices in our ODA-guided data selection and sampling framework.

\paragraph{Analysis of metric-based selection.}
To further investigate the effectiveness of existing data curation paradigms, we compare our ODA-guided selection against the common practice of metric-based scoring aggregation through our ODA-Tool\footnote{\url{https://github.com/OpenDataArena/OpenDataArena-Tool/tree/main/data_scorer}}. We employ five representative automated scorers: \textit{Deita-Complexity}~\cite{deita}, \textit{Deita-Quality}~\cite{deita}, \textit{Reward Model}~\cite{liu2025skywork}, \textit{LLM-as-Judge}, and \textit{IFD}~\cite{ifd}. Using data from the top-20 ODA datasets, we construct two 500k mixtures: one using an unweighted average of these metrics and another using an empirically tuned weighted average (assigning relative coefficients of 1.0, 0.5, 1.0, 0.5, and 0.5 to the five scorers, respectively). Results are summarized in Table~\ref{tab:weighted_or_not_comparison}. The analysis reveals two primary observations:



\begin{table*}[t!]
\centering
\caption{Performance comparison of ODA mixtures against metric-based scoring aggregations and representative open-source baselines from the ODA leaderboard on Qwen2.5-7B. `Unweighted-setting' and `Weighted-setting' denote 500k mixtures curated from the top-20 ODA datasets; the latter applies relative coefficients of 1.0, 0.5, 1.0, 0.5, and 0.5 to \textit{Deita-Complexity}, \textit{Deita-Quality}, \textit{Reward Model}, \textit{LLM-as-Judge}, and \textit{IFD}, respectively. Eff. represents Data Efficiency.}
\label{tab:weighted_or_not_comparison}
\begin{tabular}{l | c c | c c c c | c}
\toprule
\textbf{Dataset} & \textbf{Size} & \textbf{Eff.} & \textbf{General} & \textbf{Math} & \textbf{Code} & \textbf{Reasoning} & \textbf{Avg} \\
\midrule
\multicolumn{8}{c}{\textbf{Qwen2.5-7B-Base}} \\
\midrule
Qwen2.5-7B-Base~\cite{qwen2-5} & & & 51.4 & 39.8 & 50.1 & 42.7 & 46.0 \\
\midrule

SYNTHETIC-2-SFT~\cite{synthetic-2}       & 105k & \underline{+0.086} & 51.3      & 69.8          & 40.1          & \underline{58.9} & 55.0 \\
OmniThought-0528~\cite{omni-thought}              & 365k & +0.027         & 47.1          & 71.2          & 47.6          & 57.2          & 55.8 \\
Unweighted-setting & 500k & +0.024 & \underline{60.8} & 65.9 & 53.9 & 51.9 & 58.1 \\
OpenThoughts3~\cite{{open-thoughts}}          & 1.2M & +0.011         & 45.5          & \underline{71.8}          & \textbf{67.0} & 54.3          & 59.6 \\
Weighted-setting & 500k & +0.030 & 60.2 & 70.4 & 59.4 & 53.9 & 60.9 \\
\rowcolor{lightgray} ODA-Mixture-100k & 101k & \textbf{+0.149} & 56.8 & 71.2 & 64.4 & 51.5 & \underline{61.0} \\
\rowcolor{lightgray} ODA-Mixture-500k & 506k & +0.039 & \textbf{63.4} & \textbf{72.8} & \underline{66.7} & \textbf{59.6} & \textbf{65.6} \\

\bottomrule
\end{tabular}%
\end{table*}

\begin{itemize}
    \item Metric-based selection demonstrates strong inherent utility; even the unweighted average (58.1 Avg) yields performance competitive with previous top-tier benchmarks, while modest empirical tuning in the weighted version (60.9 Avg) comfortably secures a leading position. These results highlight that automated proxy metrics capture essential quality dimensions, suggesting that researching more effective, automated ways to aggregate these diverse signals remains a highly promising and fruitful direction.
    \item ODA-guided mixtures demonstrate far superior efficiency. Notably, ODA-Mixture-100k outperforms the best weighted 500k baseline using only 20\% of the data, achieving a five-fold increase in Data Efficiency (+0.149 vs. +0.030). ODA-Mixture-500k further consolidates this advantage, surpassing the best scoring baseline by a substantial 4.7-point margin.
\end{itemize}

These results suggest that ODA leaderboards provide a superior, task-grounded signal that implicitly integrates diverse quality dimensions. By leveraging this direct signal, we bypass the complexities of proxy metric engineering and achieve a more robust and scalable path for constructing high-performance SFT mixtures.

\paragraph{Impact of intra-cluster sampling.}
To understand the influence of within-cluster selection, we compare two sampling schemes: difficulty-priority sampling (using sequence length as a proxy for complexity) and random sampling. We analyze their performance across different data budgets on the ODA leaderboard, as illustrated in Figure~\ref{fig:scaling_trend_comparison}. Our results reveal two distinct scaling behaviors:

\begin{figure}[t]
    \centering
    \includegraphics[width=0.85\linewidth]{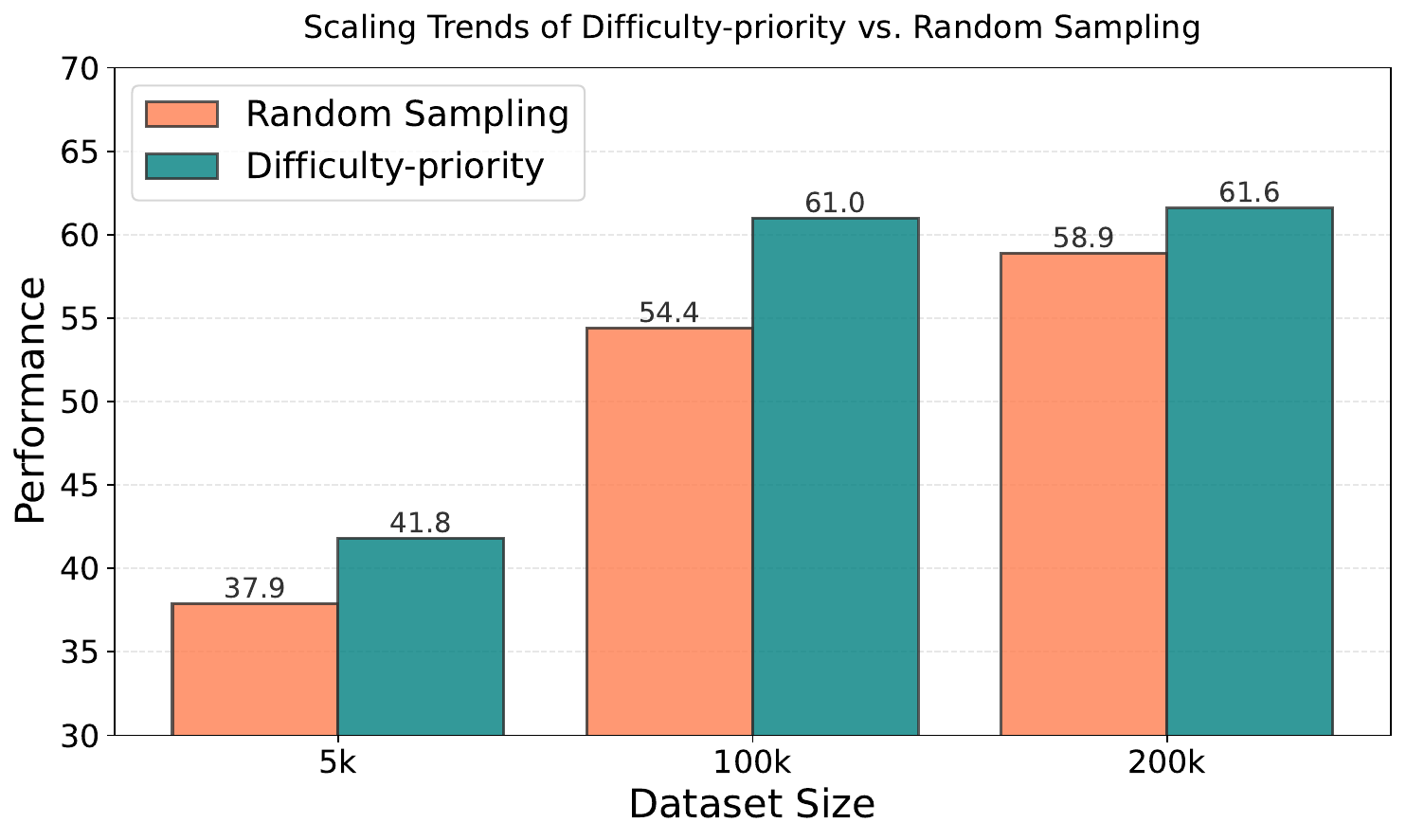}
    \caption{Scaling trends of difficulty-priority vs. random sampling within clusters.}
    \label{fig:scaling_trend_comparison}
\end{figure}

First, difficulty-priority sampling is superior in low-resource regimes. At the 100k scale, prioritizing harder instances leads to faster capability gains, particularly in Math and Code domains where base models typically struggle. Concentrating on these challenging samples allows the model to bridge critical reasoning gaps more efficiently than uniform sampling.

Second, random sampling exhibits better long-term scalability. While difficulty-priority sampling delivers strong initial growth, its marginal benefits saturate as the budget increases, likely because the selected high-difficulty data becomes increasingly homogeneous. In contrast, random sampling maintains a steadier improvement curve at larger scales. The continued introduction of diverse examples provides broader exposure to various task formats and distributions, preventing the performance plateau observed with difficulty-focused data.

These findings directly justify our two-track strategy. The effectiveness of difficulty-priority sampling at small scales motivates ODA-Mixture-100k, where efficiency is paramount. Conversely, the superior scaling of diversity-oriented sampling at larger volumes informs the design of ODA-Mixture-500k. Overall, this ablation emphasizes that optimal sampling dynamics must be tailored to the target data budget to balance rapid capability acquisition with long-term generality.

\subsection{Data Analysis}
To better understand the structural and semantic characteristics of the curated ODA-Mixtures, we conduct a set of qualitative and distributional analyses. In particular, we examine semantic coverage and sequence length distributions to complement the quantitative results.

\subsubsection{Embedding Analysis and Semantic Diversity.}
To further analyze the semantic properties of our curated mixtures, we visualize the embedding distributions of ODA-Mixture-500k and ODA-Mixture-100k in comparison with several strong and broadly applicable open-source datasets. These baselines include Light-R1-SFTData, OmniThought, OpenThoughts3, SYNTHETIC-2-SFT-verified, and MegaScience, all of which are comprehensive datasets and perform competitively on the \texttt{Qwen2.5} and \texttt{Qwen3} leaderboards. For each dataset, we randomly sample 50k instances, generate embeddings using the \texttt{Qwen3-Embedding-8B} model~\cite{qwen3}, and project them into two dimensions using t-SNE~\cite{van2008tsne} for visualization.

\begin{figure}[htbp]
  \centering

  \begin{subfigure}[t]{0.48\textwidth}
    \centering
    \includegraphics[width=\linewidth]{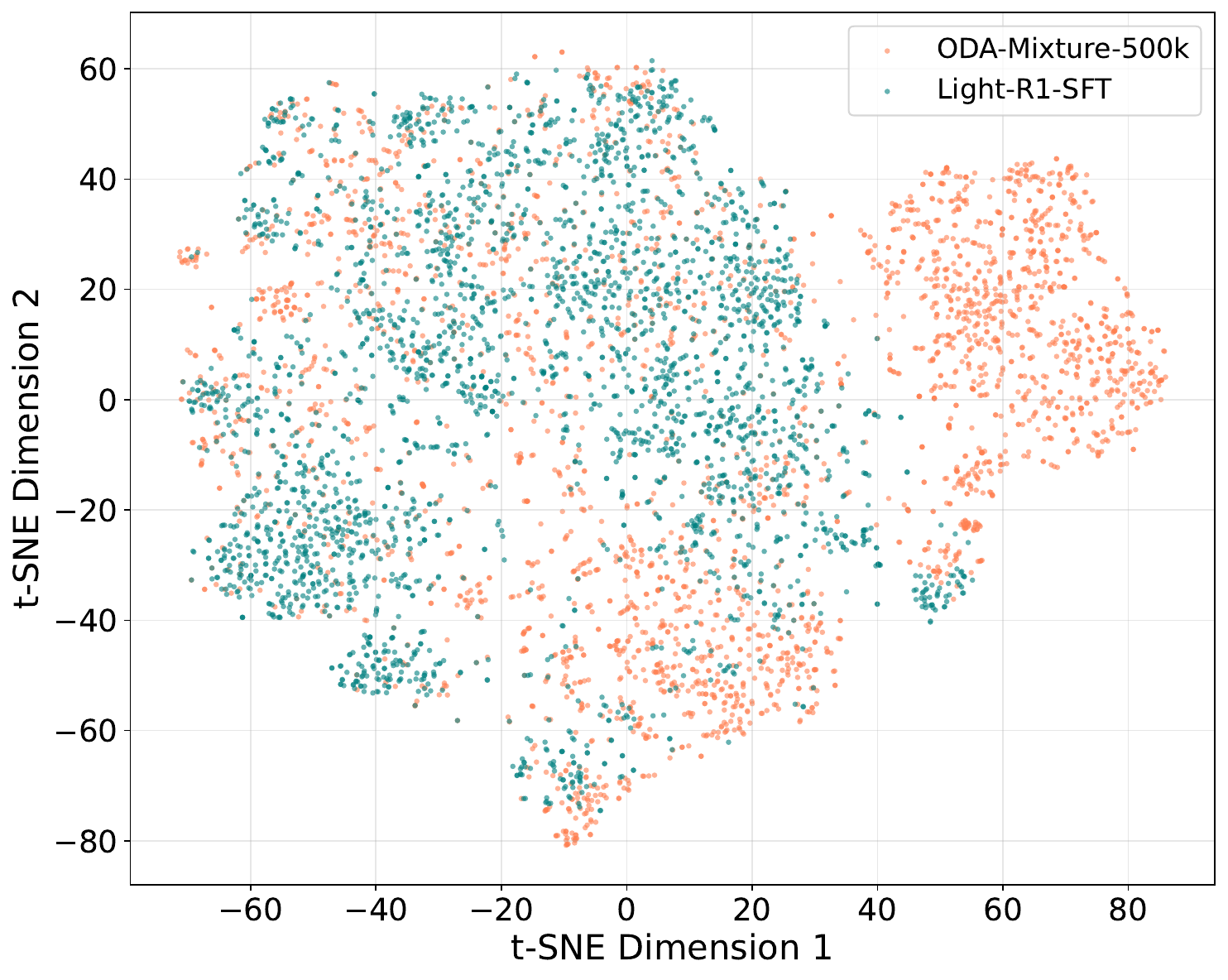}
    \caption{Embedding comparison between ODA-Mixture-500k and Light-R1-SFTData}
  \end{subfigure}\hfill
  \begin{subfigure}[t]{0.48\textwidth}
    \centering
    \includegraphics[width=\linewidth]{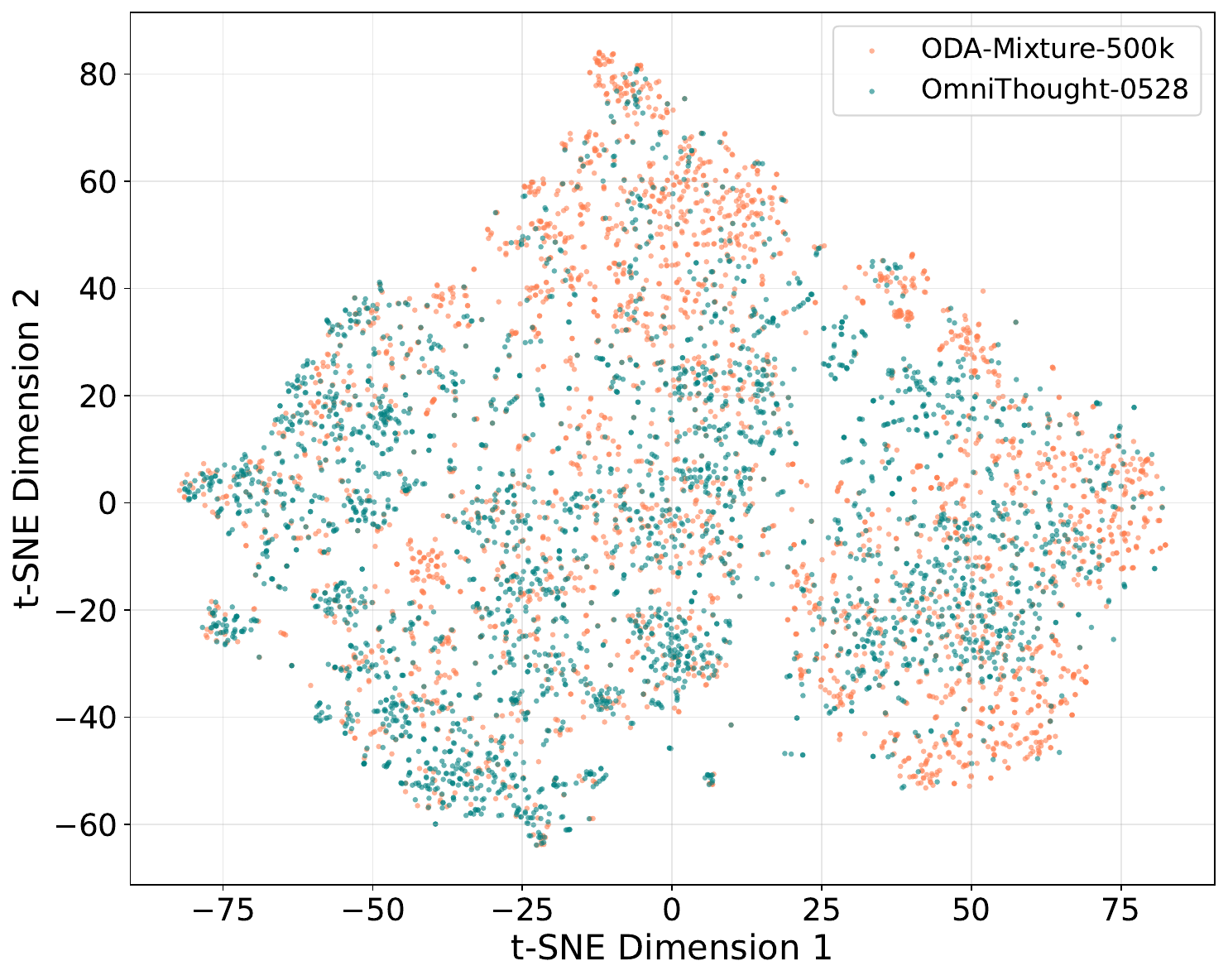}
    \caption{Embedding comparison between ODA-Mixture-500k and OmniThought}
  \end{subfigure}

  \vspace{0.8em}

  \begin{subfigure}[t]{0.48\textwidth}
    \centering
    \includegraphics[width=\linewidth]{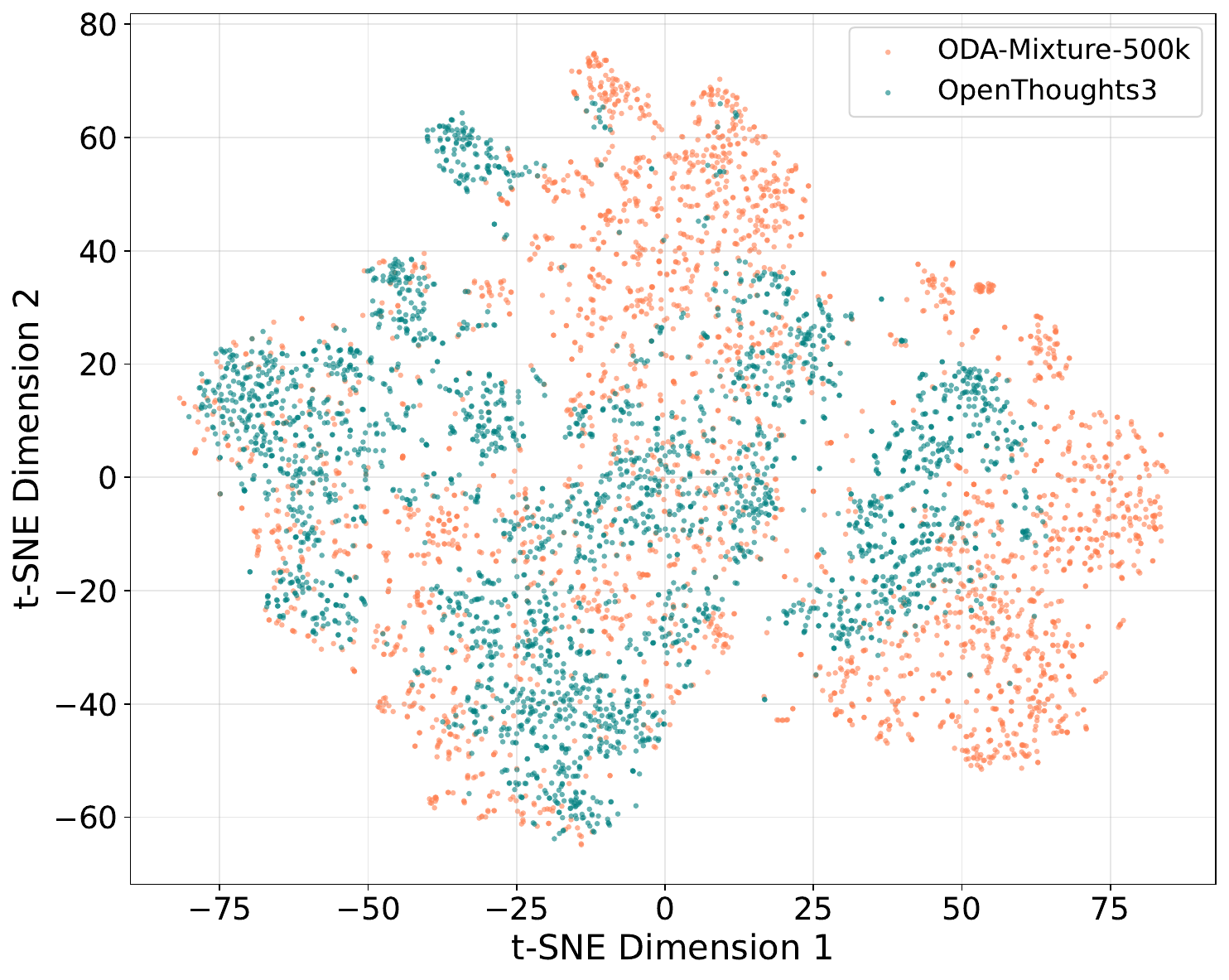}
    \caption{Embedding comparison between ODA-Mixture-500k and OpenThoughts3}
  \end{subfigure}\hfill
  \begin{subfigure}[t]{0.48\textwidth}
    \centering
    \includegraphics[width=\linewidth]{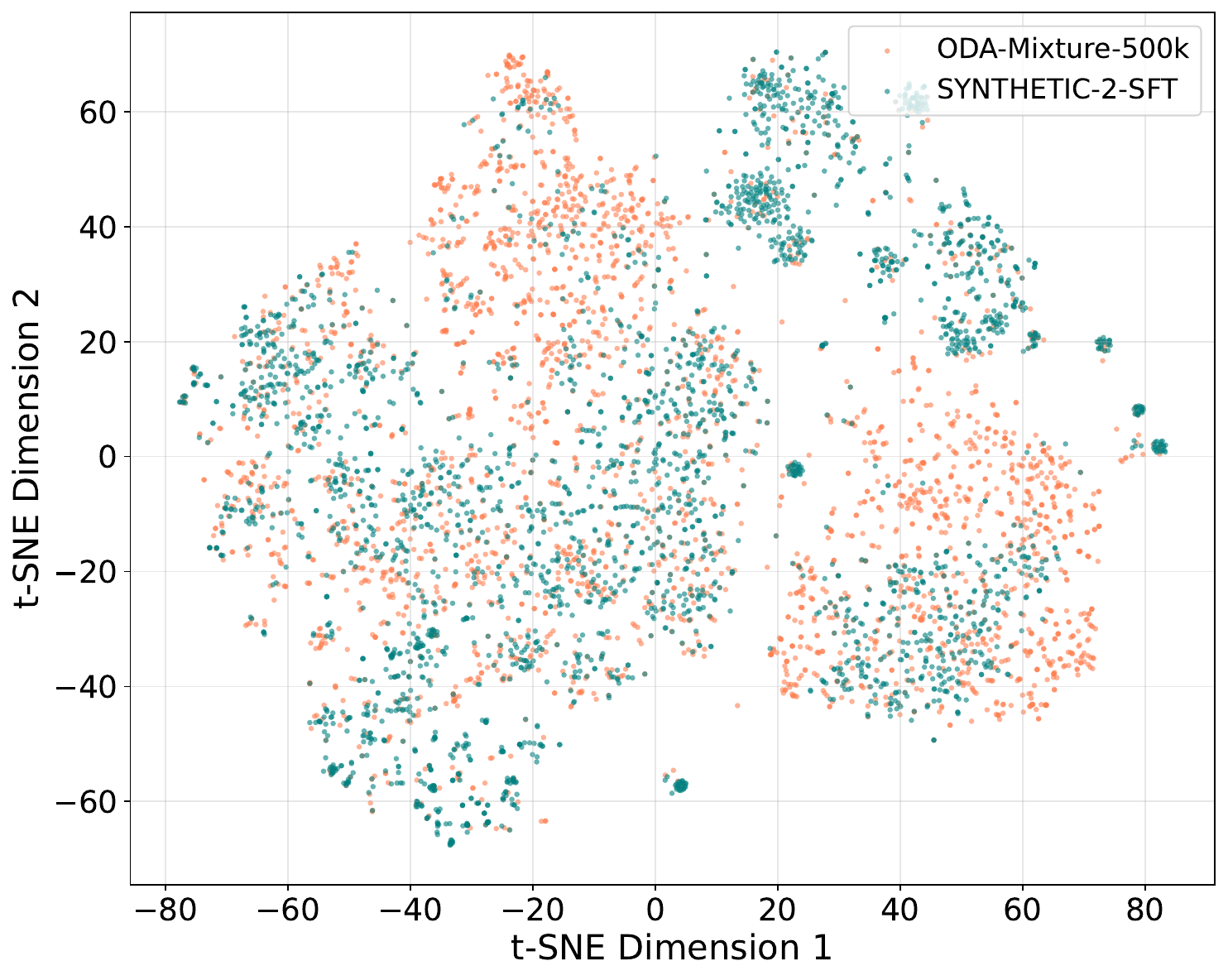}
    \caption{Embedding comparison between ODA-Mixture-500k and SYNTHETIC-2-SFT-verified}
  \end{subfigure}

  \vspace{0.8em}

  \begin{subfigure}[t]{0.48\textwidth}
    \centering
    \includegraphics[width=\linewidth]{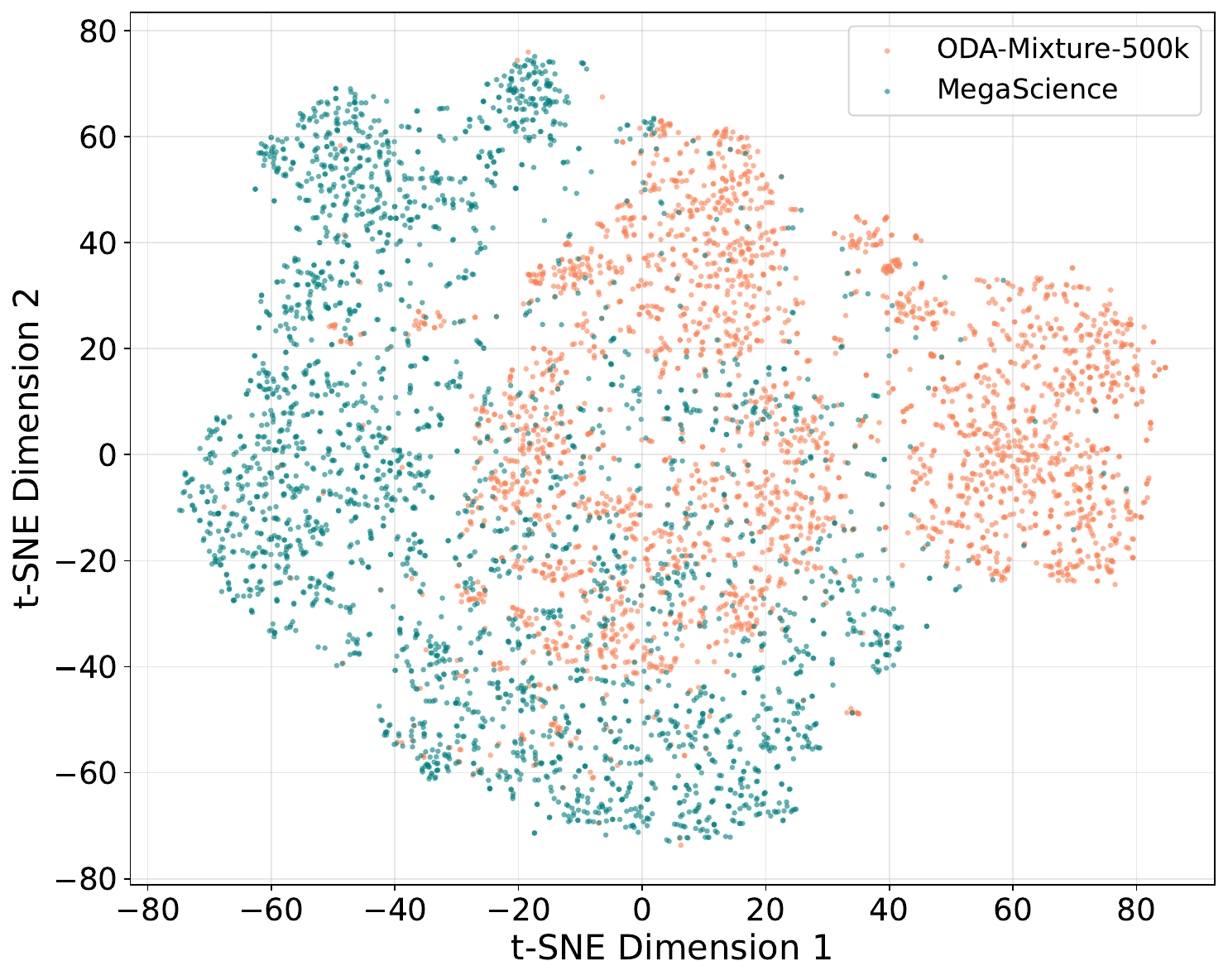}
    \caption{Embedding comparison between ODA-Mixture-500k and MegaScience}
  \end{subfigure}\hfill
  \begin{subfigure}[t]{0.48\textwidth}
    \centering
    \includegraphics[width=\linewidth]{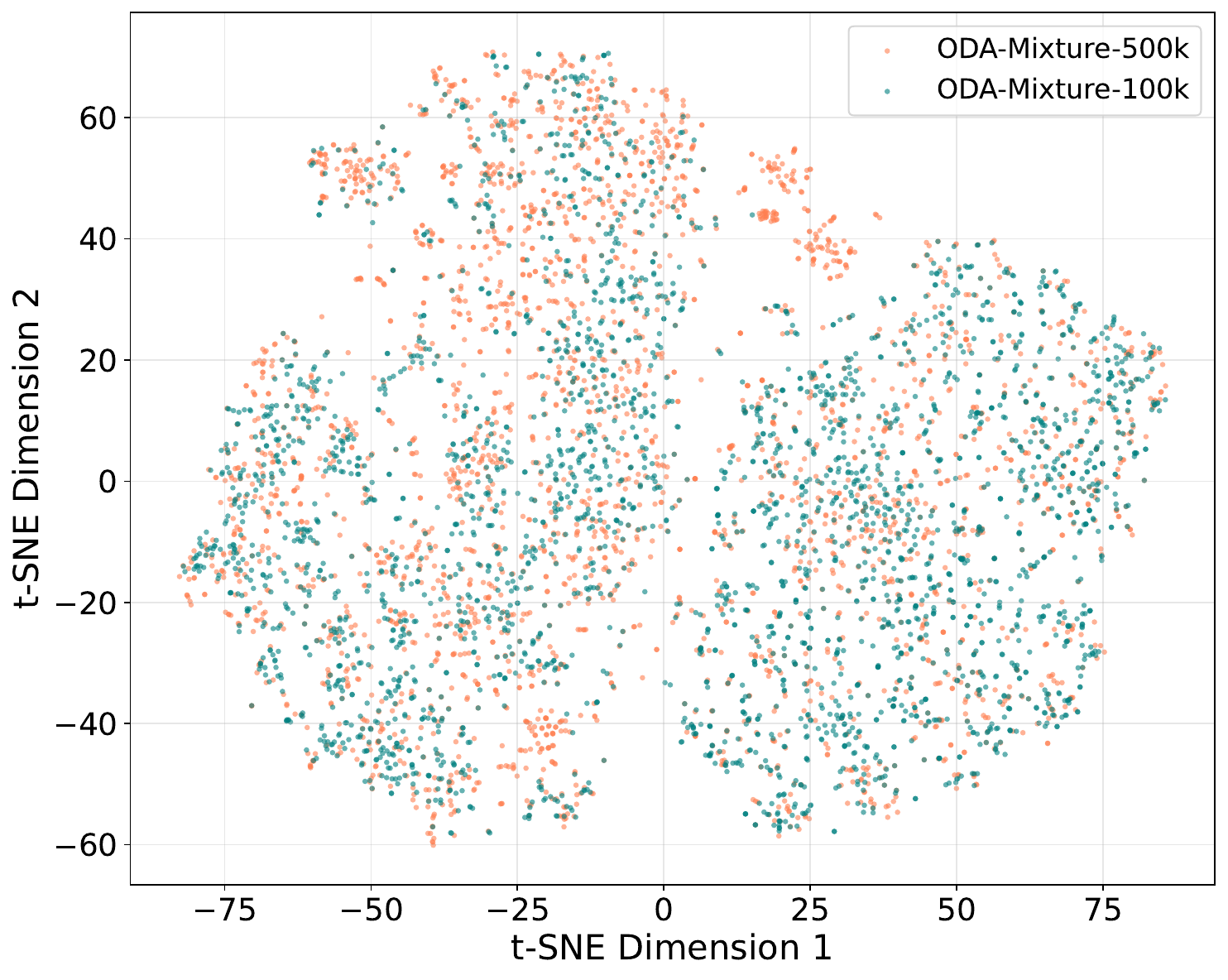}
    \caption{Embedding comparison between ODA-Mixture-500k and ODA-Mixture-100k}
  \end{subfigure}

   \caption{t-SNE visualizations of embedding distributions for ODA-Mixture-500k compared with representative baseline datasets and ODA-Mixture-100k, based on 50k randomly sampled instances per dataset.}
  \label{fig:embedding_comparison}
\end{figure}

\paragraph{Semantic coverage of ODA-Mixture-500k.}
As shown in Figure~\ref{fig:embedding_comparison}, ODA-Mixture-500k exhibits a consistently broader and more evenly distributed coverage of the semantic space compared to several widely used baselines, including Light-R1-SFTData, OmniThought, OpenThoughts3, and SYNTHETIC-2-SFT-verified. While these datasets tend to form denser clusters concentrated around specific regions of the manifold, ODA-Mixture-500k more extensively occupies both core regions and interstitial areas between clusters. This pattern suggests that our diversity-oriented mixture construction captures a wider range of instruction intents, reasoning styles, and knowledge domains.

\paragraph{Comparison with MegaScience.}
When compared with MegaScience, the difference in overall diversity is less pronounced. Both datasets exhibit substantial semantic coverage, and each contains regions of the embedding space that are not populated by the other. This observation is consistent with the design of MegaScience, which—despite being a broad dataset—places a strong emphasis on science-related content and therefore exhibits a more domain-specific distribution. In contrast, ODA-Mixture-500k covers a wider variety of non-scientific reasoning and instruction patterns. The complementary nature of their distributions further supports the conclusion that ODA-Mixture-500k is semantically diverse, even if it does not strictly dominate MegaScience along all dimensions.

\paragraph{ODA-Mixture-100k vs.\ ODA-Mixture-500k.}
Interestingly, we find that ODA-Mixture-100k also demonstrates a high degree of semantic diversity and does not exhibit a clear disadvantage relative to ODA-Mixture-500k in the embedding space. Despite being constructed under a significantly smaller data budget and with a difficulty-priority sampling strategy, the 100k mixture already spans many of the semantic regions covered by the larger mixture. This result suggests that ODA-guided data selection can achieve substantial semantic coverage even at relatively small scales, and that increasing data volume primarily densifies existing regions rather than introducing entirely new ones.

Therefore, these embedding-based analyses complement our quantitative results and provide qualitative evidence that the ODA leaderboard serves as an effective signal for constructing semantically diverse and well-balanced mixture datasets.

\subsubsection{Token Length Distribution}
We analyze the token length distributions of our curated mixtures alongside representative baselines to understand how response length---specifically its long-tail behavior---relates to model performance. As shown in Figure~\ref{fig:mix_length_distribution}, we can see:

\begin{figure}[t]
    \centering
    \includegraphics[width=0.85\linewidth]{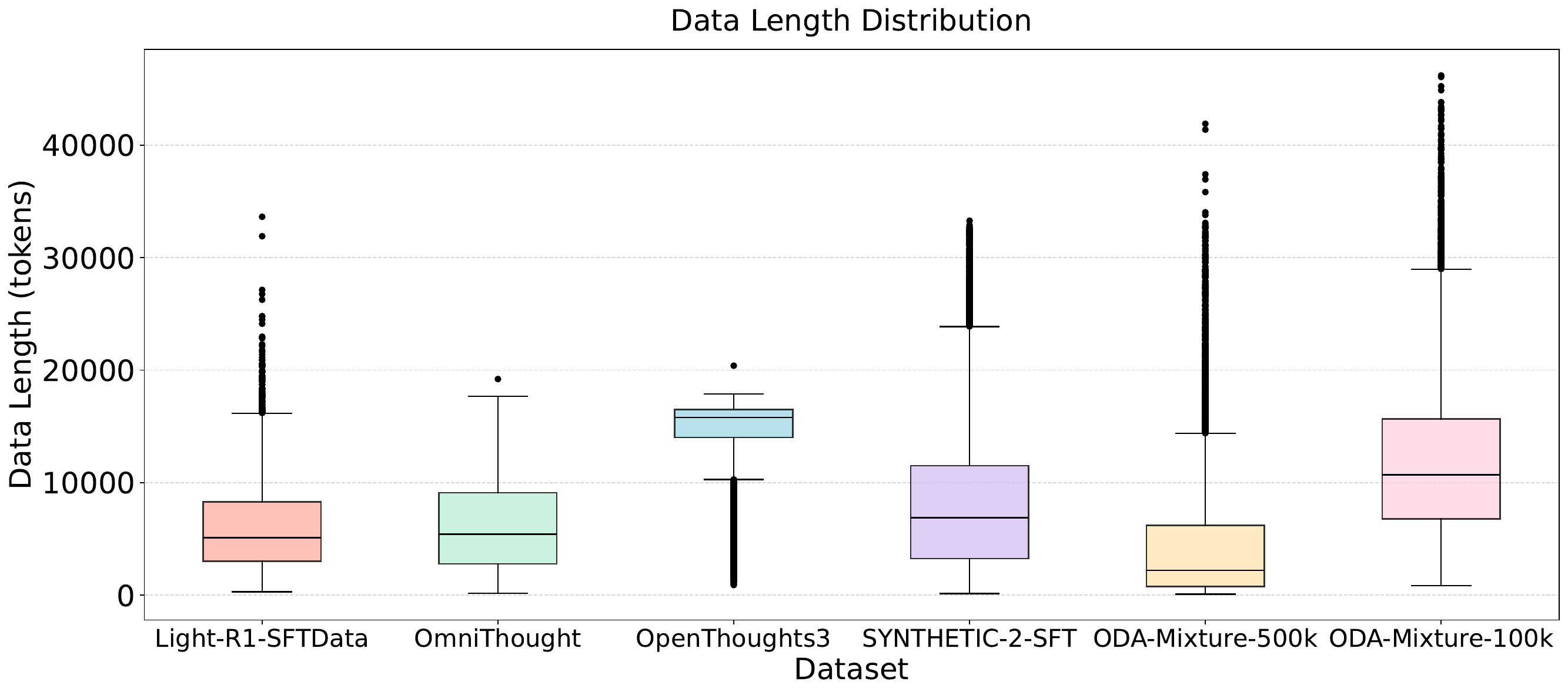}
    \caption{Token length distributions across datasets. }
    \label{fig:mix_length_distribution}
\end{figure}

\begin{itemize}
    \item \textbf{ODA-Mixtures benefit from heavy-tailed complexity.} Across the evaluated datasets, ODA mixtures exhibit a distinctive heavy-tailed profile compared to typical baselines. While datasets such as Light-R1-SFTData and OmniThought are primarily composed of short-to-medium sequences, ODA-Mixture-100k and 500k possess a significantly higher median length and a dense concentration of extreme outliers. These long sequences, primarily drawn from challenging Math and Code domains, serve as a reliable proxy for reasoning depth. By prioritizing these semantically dense samples, ODA mixtures effectively expose the model to complex compositional structures even at a limited data scale.
    \item \textbf{Selection strategies shape distinct length profiles.} The shift in length distribution reflects the different selection objectives of our two tracks. In the efficiency-oriented ODA-Mixture-100k, the use of difficulty-priority sampling deliberately maintains a high median length and a substantial long tail to maximize capability gains under tight constraints. In contrast, the diversity-oriented sampling in ODA-Mixture-500k produces a more balanced hierarchy. While it retains the essential long tail for reasoning, the distribution transitions toward medium-length samples to ensure broader coverage.
\end{itemize}


Overall, these results underscore that long-tail outliers are high-impact signals essential for reasoning performance rather than noise. By strategically preserving these samples, ODA effectively balances reasoning depth with task breadth, delivering both extreme data efficiency and robust scalability.

\section{Conclusion}

In this work, we close the loop between \emph{measuring} dataset value and \emph{engineering} better training corpora. Building on OpenDataArena (ODA) as a standardized, reproducible platform for post-training data evaluation, we show that ODA’s leaderboard signals can be used as an actionable objective for data construction—turning evaluation into feedback and transforming dataset design from ad-hoc aggregation into a principled, iterative process.

Following this paradigm, we introduce two datasets that instantiate ODA-guided engineering in complementary regimes. \textbf{ODA-Math-460k} leverages ODA’s domain evidence to aggregate high-value math sources, then applies strict deduplication and benchmark decontamination, difficulty-banded selection to retain learnable-but-challenging problems, and a synthesize-and-verify distillation pipeline with verifier-backed correctness. \textbf{ODA-Mixture} uses ODA’s overall ranking as a global utility signal and adopts an anchor-and-patch strategy to build compact yet strong multi-domain mixtures, further exploring two practical operating points: a small-budget, difficulty-prioritized efficiency track and a larger-budget, diversity-oriented performance track. Across both settings and multiple base models, the resulting corpora deliver SOTA performance and data efficiency relative to strong open baselines, demonstrating that ODA is not merely a benchmarking tool but a key enabler for constructing superior SFT data.

Looking forward, ODA opens a path toward continuous, community-driven improvement of post-training corpora: expanding coverage to more domains and languages, strengthening contamination defenses as benchmarks evolve, improving automated verification beyond final-answer matching, and optimizing mixtures under explicit compute/data budgets. More broadly, we hope this work encourages a shift toward \emph{data-centric post-training}, where transparent evaluation and iterative dataset engineering form a sustainable loop for advancing open LLM capabilities.


\clearpage
\newpage
\bibliographystyle{plainnat}
\setcitestyle{numbers}
\bibliography{paper_updated}

\clearpage
\newpage
\beginappendix

\section{Dataset List}

This section details the data composition of the ODA series, including ODA-Math-460k, ODA-Mixture-100k and ODA-Mixture-500k. The curation process began with a broad survey of high-quality resources released in 2024 and 2025.  Table~\ref{tab:full_dataset_list} and Table~\ref{tab:mix_full_dataset_list} enumerates the full pool of candidate datasets, categorized by their release year and ranked by the total number of available QA samples.

From these initial corpora, we performed rigorous performance-driven filtering to assemble our final training sets. The resulting source distributions for ODA-Math-460k are provided in Table~\ref{tab:source_dist_booktabs}, while the compositions for the ODA-Mixture-100k and ODA-Mixture-500k are detailed in Tables~\ref{tab:mix_100k_source_dist_booktabs} and \ref{tab:mix_500k_source_dist_booktabs}. These tables highlight the proportional contributions of various high-quality sources to our final data mixtures.

\begin{table}[h]
\caption{All dataset candidates for the creation of ODA-Math-460k, grouped by release year and sorted by size within each year.}
\centering
\begin{tabular}{lcr}
\toprule
\textbf{Dataset Name} & \textbf{Release Year} &  \textbf{\# QA Samples} \\
\midrule
OpenMathReasoning (CoT)~\cite{open-math-reasoning}        & 2025 & 3,201,061 \\
OpenThoughts3 (math)~\cite{open-thoughts}                 & 2025 &   850,000 \\
MiroMind-M1-SFT-719K~\cite{miro-mind}                     & 2025 &   719,232 \\
AM-Thinking-v1-Distilled (math)~\cite{am-thinking-v1-distilled} & 2025 &   558,129 \\
OpenR1-Math-220k~\cite{openr1-math-220k}                 & 2025 &   450,258 \\
MegaScience (math)~\cite{mega-science}                    & 2025 &   413,842 \\
OmniThought-0528~\cite{omni-thought}                      & 2025 &   364,988 \\
DeepMath-103K~\cite{deep-math-103k}                       & 2025 &   309,066 \\
SCP-116K~\cite{scp-116k}                                  & 2025 &   157,210 \\
Light-R1-SFT~\cite{light-r1}                              & 2025 &    79,439 \\
MiroMind-M1-RL-62K~\cite{miro-mind}                       & 2025 &    62,118 \\
MathFusionQA~\cite{math-fusion}                           & 2025 &    59,892 \\
SYNTHETIC-2-SFT-verified (math)~\cite{synthetic-2}        & 2025 &    49,781 \\
AceReason-Math~\cite{ace-reason}                          & 2025 &    49,585 \\
Omega-Problems~\cite{omega-problems}                      & 2025 &    18,885 \\
Fast-Math-R1-SFT~\cite{fast-math-r1}                      & 2025 &     7,900 \\
LIMO~\cite{limo}                                          & 2025 &       817 \\
\midrule
OpenMathInstruct-2~\cite{openmath-instruct2}              & 2024 &21,972,791 \\
ScaleQuest-Math~\cite{scale-quest-math}                   & 2024 & 1,003,467 \\
NuminaMath-1.5~\cite{numina-math}                         & 2024 &   896,215 \\
NuminaMath-CoT~\cite{numina-math}                         & 2024 &   859,594 \\
DART-Math-hard~\cite{dart-math}                           & 2024 &   585,392 \\
Magpie-Reasoning-V2-250K~\cite{magpie}                    & 2024 &   249,922 \\
math-gpt-4o-200k~\cite{math-gpt4o-200k}                   & 2024 &   200,035 \\
Magpie-Reasoning-V1-150K~\cite{magpie}                    & 2024 &   150,000 \\
\bottomrule
\end{tabular}
\label{tab:full_dataset_list}
\end{table}

\begin{table}[h]
\caption{Detailed source distribution for ODA-Math-460k.}
\centering
\begin{tabular}{l|r|r}
\toprule
\textbf{Source} & \textbf{Count} & \textbf{Percentage} \\
\midrule
ScaleQuest-Math~\cite{scale-quest-math} & 87,755 & 19.09\% \\
NuminaMath-CoT~\cite{numina-math} & 75,971 & 16.53\% \\
OpenMathInstruct-2~\cite{openmath-instruct2} & 65,688 & 14.29\% \\
MegaScience (math)~\cite{mega-science} & 54,904 & 11.94\% \\
OpenMathReasoning~\cite{open-math-reasoning} & 49,463 & 10.76\% \\
AM-Thinking-Distilled~\cite{am-thinking-v1-distilled} & 38,375 & 8.35\% \\
MiroMind-M1-SFT-719K~\cite{miro-mind} & 23,417 & 5.09\% \\
SCP-116K~\cite{scp-116k} & 16,066 & 3.50\% \\
DeepMath-103K~\cite{deep-math-103k} & 11,956 & 2.60\% \\
math-gpt-4o-200k~\cite{math-gpt4o-200k} & 8,355 & 1.82\% \\
OpenR1-Math-220k~\cite{openr1-math-220k} & 7,999 & 1.74\% \\
MathFusionQA~\cite{math-fusion} & 6,510 & 1.42\% \\
NuminaMath-1.5~\cite{numina-math} & 4,529 & 0.99\% \\
Omega-Problems~\cite{omega-problems} & 3,436 & 0.75\% \\
OmniThought-0528~\cite{omni-thought} & 2,303 & 0.50\% \\
MiroMind-M1-RL-62K~\cite{miro-mind} & 1,229 & 0.27\% \\
AceReason-Math~\cite{ace-reason} & 957 & 0.21\% \\
Fast-Math-R1-SFT~\cite{fast-math-r1} & 397 & 0.09\% \\
SYNTHETIC-2-SFT-verified (math)~\cite{synthetic-2} & 179 & 0.04\% \\
DART-Math-hard~\cite{dart-math} & 157 & 0.03\% \\
\bottomrule
\end{tabular}
\label{tab:source_dist_booktabs}
\end{table}

\begin{table}[h]
\caption{All dataset candidates formed by the anchor and patch datasets for the creation of ODA-Mixtrue sorted by size.}
\centering
\begin{tabular}{lcr}
\toprule
\textbf{Dataset Name} & \textbf{Release Year} &  \textbf{\# QA Samples} \\
\midrule

AM-Thinking-v1-Distilled (math)~\cite{am-thinking-v1-distilled} & 2025 &   558,129 \\
AM-Thinking-v1-Distilled (code)~\cite{am-thinking-v1-distilled} & 2025 &   323,965 \\
SYNTHETIC-2-SFT~\cite{synthetic-2}        & 2025 &    104,913 \\
LIMO~\cite{limo}                                          & 2025 &       817 \\
\midrule
math-gpt-4o-200k~\cite{math-gpt4o-200k}                   & 2024 &   200,035 \\
\bottomrule
\end{tabular}
\label{tab:mix_full_dataset_list}
\end{table}

\begin{table}[h]
\caption{Detailed source distribution for ODA-Mixture-100k.}
\centering
\begin{tabular}{l|r|r}
\toprule
\textbf{Source} & \textbf{Count} & \textbf{Percentage} \\
\midrule
AM-Thinking-v1-Distilled (math)~\cite{am-thinking-v1-distilled} & 50,244 & 49.59\% \\
AM-Thinking-v1-Distilled (code)~\cite{am-thinking-v1-distilled} & 50,245 & 49.60\% \\
LIMO~\cite{limo}     & 817 &      0.81\% \\
\bottomrule
\end{tabular}
\label{tab:mix_100k_source_dist_booktabs}
\end{table}

\begin{table}[h]
\caption{Detailed source distribution for ODA-Mixture-500k.}
\centering
\begin{tabular}{l|r|r}
\toprule
\textbf{Source} & \textbf{Count} & \textbf{Percentage} \\
\midrule
AM-Thinking-v1-Distilled (math)~\cite{am-thinking-v1-distilled} & 150,244 & 29.67\% \\
AM-Thinking-v1-Distilled (code)~\cite{am-thinking-v1-distilled} & 150,252 & 29.67\% \\
math-gpt-4o-200k~\cite{math-gpt4o-200k} & 100,138 & 19.78\% \\
SYNTHETIC-2-SFT~\cite{synthetic-2} & 104913 & 20.72\% \\
LIMO~\cite{limo}     & 817 &      0.16\% \\
\bottomrule
\end{tabular}
\label{tab:mix_500k_source_dist_booktabs}
\end{table}

\clearpage
\section{Detailed Hyperparameters}

To ensure the reproducibility of our results and provide a clear baseline for future work, we present the comprehensive configurations used across our training, inference, and evaluation pipelines. This section details the specific environmental variables and algorithmic choices that governed our experiments. Table~\ref{tab:training-hyperparameters} summarizes the supervised fine-tuning (SFT) settings, including our choice of optimizer schedules and memory-efficient kernels. Subsequently, Table~\ref{tab:inference-hyperparameters} outlines the generation parameters used to sample responses from our primary models, Qwen2.5-7B and Qwen3-8B. Finally, we provide a structured overview of our evaluation suite in Table~\ref{tab:math-eval-benchmarks}, specifying the prompts, evaluators, and metrics used for each domain.

\begin{table}[h]
\caption{Detailed training settings.}
\centering
\begin{tabular}{ll}
\toprule
\textbf{Parameter} & \textbf{Values} \\
\midrule
DeepSpeed Settings & ds\_z3\_config \\
template & default \\
cutoff\_len & 32768 \\
preprocessing\_num\_workers & 16 \\
packing & true \\
per\_device\_train\_batch\_size & 2 \\
graident\_accumulation\_step & 2 \\
learning\_rate & 5.0e-5 \\
use\_liger\_kernel & true \\
num\_train\_epochs & 3.0 \\
lr\_scheduler\_type & cosine \\
warmup\_ratio & 0.1 \\
\bottomrule
\end{tabular}
\label{tab:training-hyperparameters}
\end{table}

\begin{table}[h]
    \caption{Detailed inference settings for Qwen2.5 and Qwen3 models.}
    \centering
    \begin{tabular}{lll}
        \toprule
        \textbf{Parameter} & \textbf{Qwen2.5-7B-Base} & \textbf{Qwen3-8B-Base} \\
        \midrule        
        max-out-len & 32768 & 32768 \\
        hf-type & chat & chat \\
        inference setting & vllm\_qwen2\_5\_7b\_instruct & vllm\_qwen3\_8b\_instruct \\
        accelerator & vllm + cutoff & vllm + cutoff \\
        temperature & 0 & 0.6 \\
        top-p & - & 0.95 \\
        top-k & - & 20 \\
        \bottomrule
    \end{tabular}
    \label{tab:inference-hyperparameters}
\end{table}

\begin{table}[h!]
\centering
\caption{Detailed benchmark configurations for evaluation.}
\label{tab:math-eval-benchmarks}
\resizebox{\linewidth}{!}{%
\begin{tabular}{lllll}
\toprule
\textbf{Domain}                  & \textbf{Benchmarks}     & \textbf{Evaluator}              & \textbf{Shot} & \textbf{Metric}                                  \\ \midrule
\multirow{4}{*}{\textbf{General}}   & DROP                    & xVerify-9B-C      & 3 shot         & accuracy                                         \\
                                 & IFEval                  & IFEvaluator                     & 0 shot         & Average accuracy on all IFEval benchmarks        \\
                                 & AGIEval                 & xVerify-9B-C      & 5 shot         & accuracy                                         \\
                                 & MMLU-PRO                & xVerify-9B-C      & 5 shot         & Average accuracy on all mmlu-pro benchmarks      \\ \midrule
\multirow{5}{*}{\textbf{Math}}      & Omni-MATH               & Omni-Judge            & 0 shot         & accuracy                                         \\
                                 & OlympiadBenchMath       & xVerify-9B-C      & 0 shot         & accuracy                                         \\
                                 & GSM8K                   & xVerify-9B-C      & 0 shot         & accuracy                                         \\
                                 & MATH-500                & xVerify-9B-C      & 0 shot         & accuracy                                         \\
                                 & AIME'24              & xVerify-9B-C      & 0 shot         & Average accuracy of 8 run
                                 \\
                                 & AIME'25              & CompassVerifier-7B      & 0 shot         & Average accuracy of 8 run
                                 \\
                                 & HMMT-Feb'25              & CompassVerifier-7B      & 0 shot         & Average accuracy of 8 run
                                 \\
                                 & CMIMC'25              & CompassVerifier-7B      & 0 shot         & Average accuracy of 8 run
                                 \\
                                 & BRUMO'25              & CompassVerifier-7B      & 0 shot         & Average accuracy of 8 run
                                \\ \midrule
\multirow{4}{*}{\textbf{Code}}      & HumanEval               & HumanEvalEvaluator              & 0 shot         & pass@1                                           \\
                                 & HumanEval+              & HumanEvalPlusEvaluator          & 0 shot         & pass@1                                           \\
                                 & MBPP                    & MBPPEvaluator                   & 3 shot         & pass@1                                           \\
                                 & LiveCodeBench(v5)       & LCBCGgenerationEvaluator      & 0 shot         & pass@1                                           \\ \midrule
\multirow{5}{*}{\textbf{Reasoning}} & ARC\_c                  & xVerify-9B-C      & 0 shot         & accuracy                                         \\
                                 & BBH                     & xVerify-9B-C      & 0 shot         & accuracy                                         \\
                                 & KOR-Bench               & xVerify-9B-C      & 0 shot         & Average accuracy on all kor-bench benchmarks      \\
                                 & CaLM                    & CaLMEvaluator                   & 0 shot         & Average accuracy on all calm benchmarks          \\
                                 & GPQA                    & xVerify-9B-C      & 0 shot         & accuracy                                         \\ \bottomrule
\end{tabular}%
}
\end{table}

\clearpage

\section{Definition of Data Efficiency}
\label{sec:Appendix_Data_Efficiency}

In this appendix, we formally define the Data Efficiency metric used throughout the paper to characterize the cost-effectiveness of post-training datasets. Data Efficiency measures the performance improvement obtained per unit of fine-tuning data.

\begin{equation}
    DE_{i,M} = \frac{S^{\mathrm{SFT}}_{i,M} - S^{\mathrm{Base}}_{M}}{|D_i|},
\end{equation}

where $|D_i|$ denotes the size of dataset $D_i$, $S^{\mathrm{Base}}_{M}$ is the performance of the base model $M$, and $S^{\mathrm{SFT}}_{i,M}$ is the performance of model $M$ after supervised fine-tuning on $D_i$. 

Intuitively, Data Efficiency quantifies the value density of a dataset, indicating how much performance gain can be achieved per unit of training data. This metric enables a normalized comparison across datasets of different scales and is particularly useful for analyzing efficiency under constrained data or compute budgets.

\section{Scoring Metrics}
\label{sec:Appendix_Scorers}

Here, we briefly introduce the scoring metrics used for data filtering as discussed in Section~\ref{Ablation_Studies}.

\paragraph{Deita-Complexity.}
Deita-Complexity~\cite{deita} estimates the instruction-following difficulty by predicting how cognitively demanding an instruction is for a model to execute. Higher scores indicate more complex instructions that require stronger reasoning or compositional capabilities.

\paragraph{Deita-Quality.}
Deita-Quality~\cite{deita} evaluates the overall quality of instruction--response pairs, focusing on clarity, correctness, and usefulness of the response. Higher scores correspond to higher-quality supervision signals suitable for efficient alignment.

\paragraph{Reward Model.}
We employ the \texttt{Skywork-Reward-V2-Llama-3.1-8B-40M} model~\cite{liu2025skywork} as a reward scorer to assign scalar preference scores to instruction--response pairs. Higher reward scores indicate better alignment, response quality, and instruction adherence.

\paragraph{LLM-as-Judge.}
The LLM-as-Judge framework uses \texttt{gpt-4.1-nano}~\cite{openai_gpt4} as an automated evaluator to assess multiple attributes of instruction--response pairs, including relevance, correctness, coherence, completeness, clarity, and meaningfulness, following the prompt specification in Appendix~\ref{pmt:llm_judge}.

\paragraph{Instruction Following Difficulty (IFD).}
Instruction Following Difficulty (IFD)~\cite{ifd} measures how much an instruction increases generation difficulty by computing the ratio between conditional and unconditional perplexity. In our experiments, IFD scores are computed using \texttt{Qwen2.5-7B-Instruct}~\cite{qwen2-5}, where higher values indicate harder or less aligned instruction--response pairs.

\paragraph{Score Aggregation and Normalization.}
For LLM-as-Judge, multiple attribute-level scores (i.e., relevance, correctness, coherence, completeness, clarity, and meaningfulness) are first computed independently, and their arithmetic mean is used as the final LLM-as-Judge score for each sample. The other four metrics directly produce a single scalar score without attribute-level aggregation.

To ensure comparability across different metrics, all scores are normalized to a common scale before downstream use. In ablation studies, instead of uniformly averaging all metric scores, we further explore a weighted aggregation strategy, where Deita-Complexity, Deita-Quality, Reward Model, LLM-as-Judge, and IFD are assigned weights of 1.0, 0.5, 1.0, 0.5, and 0.5, respectively, reflecting their relative importance in data filtering.

\section{Prompts}
\captionsetup{type=prompt}
\caption{Prompt for math domain detection and subject classification.}
\label{pmt:math_classification}

\begin{tcolorbox}[
  colback=gray!5,
  colframe=black!75,
  width=\textwidth,
  title=Math Domain Classification Prompt,
  breakable
]
\small

You are a strict Math Domain Classifier. Your task is to analyze the user’s input and categorize it into one of the specific domains listed below. If the input is NOT a math problem (as defined below), you must classify it as \textbf{'Non-Math'}.

Output EXACTLY one \texttt{<answer>...</answer>} tag containing the category name (or 'Non-Math'). Do not provide explanations or extra text.

\medskip
\textbf{Supported Categories (and what they include)}

\begin{enumerate}
  \item \textbf{Algebra:} Includes Linear Algebra (matrices, vectors), Pre-Algebra, Arithmetic (basic calculations), and General Algebra (solving equations, inequalities).
  \item \textbf{Geometry:} Includes Euclidean Geometry (shapes, spatial reasoning) and Trigonometry.
  \item \textbf{Calculus:} Includes Derivatives, Integrals, Limits, Series, and Differential Equations.
  \item \textbf{Discrete \& Probability:} Includes Combinatorics, Probability, Statistics, Set Theory, Graph Theory, and Logic puzzles.
  \item \textbf{Number Theory:} Includes Prime numbers, Divisibility, Modular Arithmetic, and Diophantine equations.
  \item \textbf{Other:} Includes Applied Math (Physics/Economics word problems), Multidisciplinary problems, or math that does not fit the above.
  \item \textbf{Non-Math:} History of math, definitions without application, pure coding/syntax questions, casual conversation, or factual lookups.
\end{enumerate}

\medskip
\textbf{Classification Rules (Follow in order of priority)}

\begin{enumerate}
  \item \textbf{Filter Non-Math First:} If the input asks for a biography, a date, a definition without a problem to solve, or code syntax (e.g., ``How do I print in Python?''), output \texttt{<answer>Non-Math</answer>}.
  \item \textbf{Identify Keywords \& Context:} Look for domain-specific notation (e.g., integrals for Calculus, ``mod'' for Number Theory, matrices for Algebra).
  \item \textbf{Logic \& Counting:} If the problem involves ``how many ways'', chance, dice, or logical deduction (knights/knaves), classify as \texttt{<answer>Discrete \& Probability</answer>}.
  \item \textbf{Arithmetic \& Equations:} If the problem is purely calculation (\(2+2\)) or solving for variables (\(x^2 + y = 10\)), classify as \texttt{<answer>Algebra</answer>}.
  \item \textbf{Shapes \& Angles:} If it involves triangles, circles, or trigonometric functions (\(\sin/\cos/\tan\)), classify as \texttt{<answer>Geometry</answer>}.
  \item \textbf{Applied Contexts:} If it is a physics word problem requiring formulas (kinematics) or specific financial math, classify as \texttt{<answer>Other</answer>}.
\end{enumerate}

\medskip
\textbf{Examples}

\textbf{Example 1 (Algebra).} \\
Input: Solve for \(x\): \(3x + 5 = 20\). \\
Respond with: \texttt{<answer>Algebra</answer>}

\medskip
\textbf{Example 2 (Arithmetic \(\to\) Algebra).} \\
Input: What is \(25 \times 45\)? \\
Respond with: \texttt{<answer>Algebra</answer>}

\medskip
\textbf{Example 3 (Geometry).} \\
Input: Find the hypotenuse of a right triangle with legs 3 and 4. \\
Respond with: \texttt{<answer>Geometry</answer>}

\medskip
\textbf{Example 4 (Calculus).} \\
Input: Calculate the derivative of \(f(x) = e^x \sin(x)\). \\
Respond with: \texttt{<answer>Calculus</answer>}

\medskip
\textbf{Example 5 (Discrete \& Probability).} \\
Input: In how many distinct ways can the letters of the word MISSISSIPPI be arranged? \\
Respond with: \texttt{<answer>Discrete \& Probability</answer>}

\medskip
\textbf{Example 6 (Number Theory).} \\
Input: Find the remainder when \(2^{100}\) is divided by 7. \\
Respond with: \texttt{<answer>Number Theory</answer>}

\medskip
\textbf{Example 7 (Non-Math - History).} \\
Input: Who is known as the Prince of Mathematicians? \\
Respond with: \texttt{<answer>Non-Math</answer>}

\medskip
\textbf{Example 8 (Non-Math - Coding).} \\
Input: Write a React component to display a button. \\
Respond with: \texttt{<answer>Non-Math</answer>}

\medskip
\textbf{Example 9 (Other - Applied).} \\
Input: A car accelerates at \(5 \text{ m/s}^2\) for 10 seconds. How far does it travel? \\
Respond with: \texttt{<answer>Other</answer>}

\medskip
\textbf{Task}

Input: \\
\texttt{\{instruction\}}

Respond with: \texttt{<answer>...</answer>}

\end{tcolorbox}

\captionsetup{type=prompt}
\caption{Prompt for math problem validation.}
\label{pmt:math_validation}

\begin{tcolorbox}[
  colback=gray!5,
  colframe=black!75,
  width=\textwidth,
  title=Math Problem Validation Prompt,
  breakable
]
\small

You are a rigorous math problem validator. Your task is to read the user’s input and determine if it constitutes a valid, complete, and solvable math problem based \textbf{strictly} on the provided text.

Output EXACTLY one \texttt{<answer>...</answer>} tag containing either \texttt{'YES'} or \texttt{'NO'}, with no extra text or explanations.

\medskip
\textbf{Validation Rules (Follow in order of priority)}

\begin{enumerate}
  \item \textbf{Completeness.} The problem must contain all necessary information, variables, and constraints required to solve it. If a critical value is missing (e.g., ``Solve for \(x\) where \(x + y = 10\)'' but \(y\) is never defined), output \texttt{<answer>NO</answer>}.

  \item \textbf{Logical Consistency.} The problem must not contain obvious contradictions that make it impossible (e.g., ``A triangle has side lengths 1, 1, and 10'' or ``\(x > 5\) and \(x < 2\)''). If contradictory, output \texttt{<answer>NO</answer>}.

  \item \textbf{Text Integrity.} The text must be readable. If the problem is cut off mid-sentence, contains severe encoding garbage that obscures the numbers, or ends abruptly without a question, output \texttt{<answer>NO</answer>}.

  \item \textbf{Ambiguity.} The goal must be clear. If the problem is ``\(x^2 + 2x\)'' with no instruction (like ``factor'', ``solve for 0'', or ``graph''), output \texttt{<answer>NO</answer>}.

  \item \textbf{Solvability (Yes).} If the problem is complete, self-contained, and logically sound---even if the math is extremely difficult or requires standard context assumptions (e.g., standard gravity, base-10 system)---output \texttt{<answer>YES</answer>}.

  \item \textbf{Formatting Tolerance.} Do NOT mark a problem as invalid solely due to minor LaTeX syntax errors (e.g., missing a closing brace) provided the mathematical intent remains unambiguous.
\end{enumerate}

\medskip
\textbf{Examples}

\textbf{Example 1 (Valid Problem).} \\
Problem: If \(f(x) = x^2 + 2\), find \(f(3)\). \\
Respond with: \texttt{<answer>YES</answer>}

\medskip
\textbf{Example 2 (Missing Information).} \\
Problem: A store sells apples for \$0.50. How much does Alice spend? \\
Respond with: \texttt{<answer>NO</answer>}

\medskip
\textbf{Example 3 (Ambiguous/No Question).} \\
Problem: Given that \(x = 5\) and \(y = 10\). \\
Respond with: \texttt{<answer>NO</answer>}

\medskip
\textbf{Example 4 (Contradictory).} \\
Problem: Find an integer \(x\) such that \(x\) is even and \(x\) is odd. \\
Respond with: \texttt{<answer>NO</answer>}

\medskip
\textbf{Task}

Math problem: \\
\texttt{\{instruction\}}

Respond with: \texttt{<answer>...</answer>}

\end{tcolorbox}

\captionsetup{type=prompt}
\caption{Prompt for solution check and answer extraction.}
\label{pmt:answer_extraction}

\begin{tcolorbox}[
  colback=gray!5,
  colframe=black!75,
  width=\textwidth,
  title=Answer Extraction Prompt,
  breakable
]
\small

You are a precise math answer extractor. Your task is to read the user’s question and the provided solution, then extract ONLY the final answer(s).

Output EXACTLY one \texttt{<answer>...</answer>} tag containing only the final answer, with no extra text or explanations.

\medskip
\textbf{Extraction Rules (Follow in order of priority)}

\begin{enumerate}
  \item \textbf{Top Priority (\(\boxed{\cdots}\)).} If a final \(\boxed{\cdots}\) is present, output its INNER CONTENT EXACTLY as written, preserving all LaTeX, symbols, and text. This rule takes precedence over all other rules (including the unit rule). Ensure the extracted content is complete (e.g., balanced braces).

  \item \textbf{Final Result (No Box).} If no \(\boxed{\cdots}\) is found, extract the final explicit numerical or symbolic result (e.g., after ``final answer is'', ``answer is'', ``Thus'', ``Therefore'').

  \item \textbf{LaTeX Preservation.} When applying Rule 2, preserve all LaTeX expressions and symbols (e.g., \verb|\sqrt{...}|, \(\infty\), \(\frac{\cdots}{\cdots}\), \(\pi\)). Do NOT convert LaTeX to plain numbers or words.

  \item \textbf{No Simplification.} Do NOT convert words to digits, rewrite mixed numbers, or simplify fractions unless they already appear that way in the final result.

  \item \textbf{Unit Stripping (No Box Only).} If applying Rule 2 (i.e., no \(\boxed{\cdots}\) was found), do NOT include units (e.g., cm, dollars, ways). Exception: Always keep the percent sign (\%).

  \item \textbf{Multiple Solutions.} If the final answer lists multiple distinct values (e.g., ``\(x=5\) or \(x=10\)'', ``the roots are \(-1\) and \(1\)''), output them as a single, comma-separated string (e.g., \texttt{"5, 10"}, \texttt{"-1, 1"}).

  \item \textbf{Word Answers.} If the solution's final answer is a definitive word (e.g., ``Yes'', ``No'', ``True'', ``False'', ``None'', ``Cannot be determined''), extract that word.

  \item \textbf{Not Found.} If no specific, concise answer (mathematical, expression, or definitive word) can be found, respond with \texttt{<answer></answer>}.
\end{enumerate}

\medskip
\textbf{Examples}

\textbf{Example 1 (Boxed).}

Solution: \texttt{...blah blah... The answer is $\boxed{10}$.} \\
Respond with: \texttt{<answer>10</answer>}

\medskip
\textbf{Example 2 (Boxed with LaTeX).}

Solution: \texttt{...so the value is $\boxed{\frac{\sqrt{3}}{2}}$.} \\
Respond with: \texttt{<answer>\textbackslash frac\{\textbackslash sqrt\{3\}\}\{2\}</answer>}

\medskip
\textbf{Example 3 (Boxed with Units --- Rule 1 Precedence).}

Solution: \texttt{...the final area is $\boxed{24 \text{ cm}^2}$.} \\
Respond with: \texttt{<answer>$24 \text\{cm\}^2$</answer>}

\medskip
\textbf{Example 4 (No Box with Units --- Rule 5 Applies).}

Solution: \texttt{...Therefore, the length is 40 meters.} \\
Respond with: \texttt{<answer>40</answer>}

\medskip
\textbf{Example 5 (No Box with Percent --- Rule 5 Exception).}

Solution: \texttt{...The total increase was 15.5\%.} \\
Respond with: \texttt{<answer>15.5\%</answer>}

\medskip
\textbf{Example 6 (Multiple Solutions).}

Solution: \texttt{...the roots of the equation are $x = -2$ or $x = 5$.} \\
Respond with: \texttt{<answer>-2, 5</answer>}

\medskip
\textbf{Example 7 (Word Answer).}

Solution: \texttt{...we can conclude that the statement is False.} \\
Respond with: \texttt{<answer>False</answer>}

\medskip
\textbf{Example 8 (Not Found).}

Solution: \texttt{...this completes the proof by induction.} \\
Respond with: \texttt{<answer></answer>}

\medskip
\textbf{Task template}

\medskip
Question: \texttt{\{instruction\}} \\[0.3em]
Solution: \texttt{\{output\_tail\}} \\[0.3em]
Respond with: \texttt{<answer>...</answer>}

\end{tcolorbox}

\captionsetup{type=prompt}
\caption{Prompt for AoPS math difficulty scoring.}
\label{pmt:math_difficulty}

\begin{tcolorbox}[
  colback=gray!5,
  colframe=black!75,
  width=\textwidth,
  title=Math Difficulty Scorer (AoPS Standard),
  breakable
]
\small

You are an experienced math olympiad coach responsible for evaluating the difficulty of a given math problem. Your goal is to assign a difficulty level from 1 to 10 based on the AoPS (Art of Problem Solving) standard.

Output EXACTLY one integer enclosed in \texttt{<score>...</score>} tags. Do not provide explanations or extra text.

\medskip
\textbf{AoPS Standard (1-10 Scale)}

The scale roughly corresponds to the USA tier system: AMC 8 $\to$ AMC 10 $\to$ AMC 12 $\to$ AIME $\to$ USA(J)MO $\to$ IMO.

\begin{enumerate}
  \item \textbf{Level 1 (Beginner):} Elementary/Middle school level (MOEMS, AMC 8 Q1-20, AMC 10 Q1-10). Standard techniques and traditional word problems.
  \item \textbf{Level 2 (Motivated Beginner):} Harder AMC 8, AMC 10 Q11-20, AIME Q1-3. Involves complex problem-solving steps.
  \item \textbf{Level 3 (Advanced Beginner):} Creative thinking required (AMC 10 Q21-25, AIME Q4-6). Harder National MATHCOUNTS instructions.
  \item \textbf{Level 4 (Intermediate):} AMC 12 Q21-25, AIME Q7-9. Complex recursion or intermediate algebra/geometry.
  \item \textbf{Level 5 (Late Intermediate):} Difficult AIME problems (Q10-12), simple proof-based Olympiad style (early JBMO, easiest USAJMO).
  \item \textbf{Level 6 (Advanced):} High-level AIME (Q13-15), Introductory Olympiad (harder USAJMO, easier USAMO/IMO Q1/4).
  \item \textbf{Level 7 (Olympiad):} Tougher Olympiad instructions requiring technical knowledge (harder USAJMO, easy-medium USAMO/IMO Q2/5).
  \item \textbf{Level 8 (High Olympiad):} Medium-hard USAMO and IMO Q2/5, easiest USAMO/IMO Q3/6.
  \item \textbf{Level 9 (Expert Olympiad):} Average USAMO and IMO Q3/6. Highly complex proofs.
  \item \textbf{Level 10 (Historic):} Historically hard problems, exceedingly tedious or difficult even for IMO standards (e.g., IMO 2020/6).
\end{enumerate}

\medskip
\textbf{Examples}

\textbf{Example (Level 1).} \\
Input: How many integer values of $x$ satisfy $|x| < 3\pi$? \\
Respond with: \texttt{<score>1</score>}

\medskip
\textbf{Example (Level 2).} \\
Input: A fair 6-sided die is rolled until an odd number appears. What is the probability that every even number appears at least once before the first odd number? \\
Respond with: \texttt{<score>2</score>}

\medskip
\textbf{Example (Level 3).} \\
Input: Triangle $ABC$ with $AB=50$ and $AC=10$ has area 120. $D$ and $E$ are midpoints of $\overline{AB}$ and $\overline{AC}$. Angle bisector of $\angle BAC$ intersects $\overline{DE}$ and $\overline{BC}$ at $F$ and $G$. Find area of $FDBG$. \\
Respond with: \texttt{<score>3</score>}

\medskip
\textbf{Example (Level 4).} \\
Input: Sequence $x_0=5$ and $x_{n+1}=\frac{x_n^2+5x_n+4}{x_n+6}$. Let $m$ be the least integer such that $x_m \leq 4+\frac{1}{2^{20}}$. In which interval does $m$ lie? \\
Respond with: \texttt{<score>4</score>}

\medskip
\textbf{Example (Level 5).} \\
Input: Find real triples $(a, b, c)$ such that $a+b+c=\frac{1}{a}+\frac{1}{b}+\frac{1}{c}$ and $a^2+b^2+c^2=\frac{1}{a^2}+\frac{1}{b^2}+\frac{1}{c^2}$. \\
Respond with: \texttt{<score>5</score>}

\medskip
\textbf{Example (Level 6).} \\
Input: Tangent to circumcircle of $\triangle HBC$ at $H$ intersects circumcircle of $\triangle ABC$ at $X, Y$. Given $HA=3, HX=2, HY=6$. Area is $m\sqrt{n}$. Find $m+n$. \\
Respond with: \texttt{<score>6</score>}

\medskip
\textbf{Example (Level 7).} \\
Input: Show that for all $n \geq 3$, there exists a balanced set of $n$ points. Determine $n$ for which a balanced centre-free set exists. \\
Respond with: \texttt{<score>7</score>}

\medskip
\textbf{Example (Level 8).} \\
Input: Coins of denomination $1/n$ sum to $99+1/2$. Prove it is possible to split into 100 groups each summing to at most 1. \\
Respond with: \texttt{<score>8</score>}

\medskip
\textbf{Example (Level 9).} \\
Input: Set $S$ of odd primes. Prove at most one way to place $S$ on a circle such that product of neighbors is $x^2+x+k$. \\
Respond with: \texttt{<score>9</score>}

\medskip
\textbf{Example (Level 10).} \\
Input: Prove existence of constant $c$ such that for $n$ points with dist $\geq 1$, there is a line $\ell$ separating $S$ with distance to $\ell$ at least $c n^{-1/3}$. \\
Respond with: \texttt{<score>10</score>}

\medskip
\textbf{Task}

Input: \\
\texttt{\{instruction\}}

Respond with: \texttt{<score>...</score>}

\end{tcolorbox}

\captionsetup{type=prompt}
\caption{Prompt for LLM-as-a-Judge scoring.}
\label{pmt:llm_judge}

\begin{tcolorbox}[
  colback=gray!5,
  colframe=black!75,
  width=\textwidth,
  title=LLM-as-a-Judge Scoring Prompt,
  breakable
]
\small

\textbf{Role}

You are a rigorous reviewer who is responsible for evaluating the quality of the \texttt{instruction-output pair}.

\medskip
\textbf{Goal}

For the given \texttt{instruction-output pair}, you need to evaluate it according to the following dimensions: \textbf{Clarity, Coherence, Completeness, Complexity, Correctness, Meaningfulness, and Relevance}.

\medskip
\textbf{Rules}

\medskip
\textbf{Clarity}

Clarity evaluates whether the instruction-output pair is clear and understandable, and whether it uses concise language and structure so that the user can easily understand it. Your score should strictly follow the rules below:

\begin{itemize}
  \item \textbf{Score 9--10 (Excellent):} The instruction is exceptionally clear, specific, and well-formulated, leaving no room for ambiguity. The output directly, accurately, and comprehensively addresses all aspects of the instruction. The language in both is precise and concise, and the output's structure is logical and easy to follow. The pair as a whole is perfectly coherent and instantly understandable.
  \item \textbf{Score 7--8 (Good):} The instruction is clear and easily understood. The output is also clear, relevant, and effectively addresses the main points of the instruction. The overall meaning of the instruction-output pair is grasped without significant effort. Minor opportunities for improvement may exist but do not hinder comprehension.
  \item \textbf{Score 5--6 (Acceptable):} The pair is generally understandable but requires some effort. Minor ambiguities, organizational issues, or slight inaccuracies may be present, though the core intent remains discernible.
  \item \textbf{Score 3--4 (Poor):} The pair is difficult to understand due to significant clarity issues, including vague instructions or poorly structured outputs that fail to directly address the instruction.
  \item \textbf{Score 1--2 (Very Poor):} The pair is unintelligible or fundamentally flawed. The instruction is nonsensical or the output is completely irrelevant or incorrect.
\end{itemize}

\medskip
\textbf{Coherence}

Coherence evaluates the linguistic and logical consistency of the instruction-output pair as a whole.

\begin{itemize}
  \item \textbf{Score 9--10 (Excellent):} The pair is perfectly coherent, logically sound, internally consistent, and forms a seamless unit.
  \item \textbf{Score 7--8 (Good):} The pair is highly coherent with only very minor logical gaps.
  \item \textbf{Score 5--6 (Acceptable):} The pair is generally coherent but contains noticeable logical gaps or minor contradictions.
  \item \textbf{Score 3--4 (Poor):} The pair has significant coherence issues, with weak logical flow or contradictions.
  \item \textbf{Score 1--2 (Very Poor):} The pair is incoherent, with no discernible logical connection between instruction and output.
\end{itemize}

\medskip
\textbf{Completeness}

Completeness evaluates how fully and accurately the output addresses the instruction.

\begin{itemize}
  \item \textbf{Score 9--10 (Excellent):} The output thoroughly and exhaustively addresses all explicit and implicit aspects of the instruction.
  \item \textbf{Score 7--8 (Good):} The output addresses all main points effectively, missing only minor details.
  \item \textbf{Score 5--6 (Acceptable):} The output covers primary aspects but lacks depth or omits secondary details.
  \item \textbf{Score 3--4 (Poor):} The output is significantly incomplete or misinterprets the instruction.
  \item \textbf{Score 1--2 (Very Poor):} The output fails to address the instruction in any meaningful way.
\end{itemize}

\medskip
\textbf{Complexity}

Complexity evaluates the depth of knowledge and reasoning required by the instruction-output pair.

\begin{itemize}
  \item \textbf{Score 9--10 (Expert-Level):} Requires deep, expert-level understanding and complex multi-step reasoning.
  \item \textbf{Score 7--8 (Advanced):} Requires significant specialized knowledge and analytical reasoning.
  \item \textbf{Score 5--6 (Intermediate):} Requires some reasoning or domain knowledge beyond simple facts.
  \item \textbf{Score 3--4 (Simple):} Involves straightforward topics and single-step reasoning.
  \item \textbf{Score 1--2 (Trivial):} Concerns extremely simple, common-sense topics.
\end{itemize}

\medskip
\textbf{Correctness}

Correctness evaluates factual accuracy and alignment between instruction and output.

\begin{itemize}
  \item \textbf{Score 9--10 (Excellent):} The pair is fully accurate, verifiable, and correctly aligned.
  \item \textbf{Score 7--8 (Good):} The pair is largely accurate with minor inaccuracies.
  \item \textbf{Score 5--6 (Acceptable):} The pair contains noticeable flaws but remains somewhat useful.
  \item \textbf{Score 3--4 (Poor):} The pair has significant factual errors or misalignment.
  \item \textbf{Score 1--2 (Very Poor):} The pair is fundamentally incorrect or misleading.
\end{itemize}

\medskip
\textbf{Meaningfulness}

Meaningfulness evaluates the value of the instruction-output pair as training data.

\begin{itemize}
  \item \textbf{Score 9--10 (Highly Meaningful):} Exceptionally valuable and informative training example.
  \item \textbf{Score 7--8 (Meaningful):} Useful and solid training example.
  \item \textbf{Score 5--6 (Moderately Meaningful):} Limited training value.
  \item \textbf{Score 3--4 (Low Meaningfulness):} Little to no training value.
  \item \textbf{Score 1--2 (Harmful):} Harmful or useless training data.
\end{itemize}

\medskip
\textbf{Relevance}

Relevance evaluates whether the output directly addresses the instruction.

\begin{itemize}
  \item \textbf{Score 9--10 (Excellent):} Fully and precisely addresses the instruction.
  \item \textbf{Score 7--8 (Good):} Addresses main points with minimal irrelevance.
  \item \textbf{Score 5--6 (Acceptable):} Partially addresses the instruction.
  \item \textbf{Score 3--4 (Poor):} Largely irrelevant or misunderstood.
  \item \textbf{Score 1--2 (Very Poor):} Completely irrelevant.
\end{itemize}

\medskip
\textbf{Output Format}

Provide your evaluation in a JSON object with keys for each dimension. Each key's value should be an integer score from 1 to 10. Do not provide any explanation.

\begin{verbatim}
{
  "Clarity": <score_integer>,
  "Coherence": <score_integer>,
  "Completeness": <score_integer>,
  "Complexity": <score_integer>,
  "Correctness": <score_integer>,
  "Meaningfulness": <score_integer>,
  "Relevance": <score_integer>
}
\end{verbatim}

\medskip
\textbf{Instruction:}

\texttt{\{instruction\}}

\medskip
\textbf{Output:}

\texttt{\{output\}}

\medskip
\textbf{Your JSON output:}

\end{tcolorbox}

\end{document}